\documentclass[10pt,twocolumn,letterpaper]{article}

\usepackage[pagenumbers]{cvpr} 

\usepackage{graphicx}
\usepackage{amsmath}
\usepackage{amssymb}
\usepackage{booktabs}
\usepackage{multirow}
\usepackage[table,xcdraw,dvipsnames]{xcolor}
\usepackage[normalem]{ulem}
\useunder{\uline}{\ul}{}
\usepackage{arydshln} 

\usepackage{array}
\newcolumntype{?}{!{\vrule width 1.5pt}}
\usepackage{makecell}

\usepackage[pagebackref,breaklinks,colorlinks,citecolor=brown]{hyperref}

\usepackage[capitalize]{cleveref}
\crefname{section}{Sec.}{Secs.}
\Crefname{section}{Section}{Sections}
\Crefname{table}{Table}{Tables}
\crefname{table}{Tab.}{Tabs.}

\def\cvprPaperID{684} 
\def\confName{CVPR}
\def\confYear{2023}

\newcommand{\statement}[1]{\noindent\textbf{#1}}

\newcommand{\rom}[1]{\uppercase\expandafter{\romannumeral #1\relax}}
\def\modelname{FAME-ViL\xspace}

\makeatletter
\renewcommand{\maketag@@@}[1]{\hbox{\m@th\normalsize\normalfont#1}}%
\makeatother

\begin{document}

\title{FAME-ViL: Multi-Tasking Vision-Language Model \\ for Heterogeneous Fashion Tasks}

\author{Xiao Han$^{1,2}$\quad Xiatian Zhu$^{1,3}$\quad Licheng Yu\quad Li Zhang$^4$\quad Yi-Zhe Song$^{1,2}$\quad Tao Xiang$^{1,2}$ \\
$^1$ CVSSP, University of Surrey \quad 
$^2$ iFlyTek-Surrey Joint Research Centre on Artificial Intelligence \\
$^3$ Surrey Institute for People-Centred Artificial Intelligence \quad
$^4$ Fudan University \\
{\tt\small \{xiao.han, xiatian.zhu, y.song, t.xiang\}@surrey.ac.uk} \\ 
{\tt\small lichengyu24@gmail.com \quad lizhangfd@fudan.edu.cn}
}

\maketitle

\begin{abstract}
In the fashion domain, there exists a variety of vision-and-language (V+L) tasks, including cross-modal retrieval, text-guided image retrieval, multi-modal classification, and image captioning.
They differ drastically in each individual input/output format and dataset size.
It has been common to design a task-specific model and fine-tune it independently from a pre-trained V+L model (\eg, CLIP).
This results in parameter inefficiency and inability to exploit inter-task relatedness. 
To address such issues, we propose a novel 
\textbf{FA}shion-focused \textbf{M}ulti-task \textbf{E}fficient learning method
for \textbf{Vi}sion-and-\textbf{L}anguage tasks (\textbf{\modelname}) in this work.
Compared with existing approaches, \modelname applies a single model for multiple heterogeneous fashion tasks, therefore being much more parameter-efficient.
It is enabled by two novel components: 
(1) a task-versatile architecture with cross-attention adapters and task-specific adapters integrated into a unified V+L model, 
and (2) a stable and effective multi-task training strategy that supports learning from heterogeneous data and prevents negative transfer. 
Extensive experiments on four fashion tasks show that our \modelname can save 61.5\% of parameters over alternatives, while significantly outperforming the conventional independently trained single-task models.
Code is available at \href{https://github.com/BrandonHanx/FAME-ViL}{https://github.com/BrandonHanx/FAME-ViL}.

\end{abstract}

\section{Introduction}
\label{sec:intro}
A variety of real-world multi-modal, particularly Vision-and-Language (V+L) tasks exist in the fashion domain, including multi-modal recognition~\cite{liao2018interpretable,ma2017towards,rostamzadeh2018fashiongen}, multi-modal retrieval~\cite{gao2020fashionbert,wu2021fashioniq} and image captioning~\cite{yang2020facad}. 
The models developed for these tasks have been applied in diverse e-commerce applications, improving product discoverability, seller-buyer engagement, and customer conversion rate after catalogue browsing.
Intrinsically, those V+L tasks are {\bf\em heterogeneous} in terms of 
(1) different input and output formats
(\eg, text-guided garment retrieval~\cite{wu2021fashioniq} and image captioning~\cite{yang2020facad} have completely different inputs and outputs); 
(2) different dataset sizes as the annotation difficulty of each task differ (\eg, the labeling effort for text-guided image retrieval is much harder than that for text-to-image retrieval~\cite{wu2021fashioniq,liu2021cirr}).
\begin{figure}[t]
\begin{center}
\includegraphics[width=\linewidth]{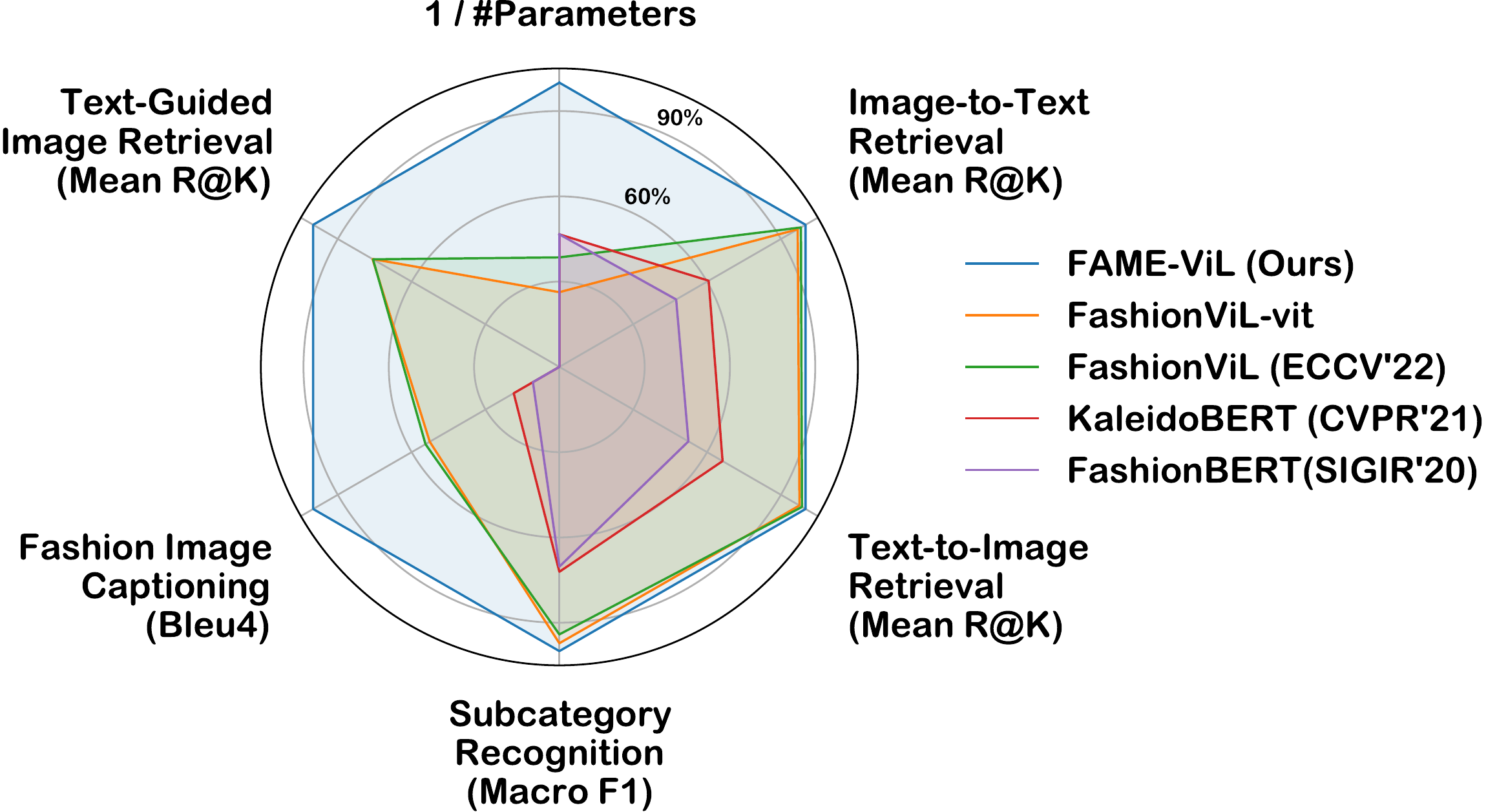}
\end{center}
\caption{By multi-task learning a single model for heterogeneous fashion tasks, our \modelname can  significantly improve parameter efficiency,
while boosting the model performance per task
over existing {\em independently fine-tuned single-task models}.
Note, each axis is \textbf{normalized} according to the respective maximum value for easier visualization.}
\label{fig:trailer}
\end{figure}

Due to the heterogeneous nature of the V+L fashion tasks, existing methods~\cite{gao2020fashionbert,zhuge2021kaleido,han2022fashionvil,yu2022commercemm,ji2022mvlt} typically take a pre-trained generic V+L model~\cite{li2019visualbert,tan2019lxmert,lu2019vilbert,su2019vlbert,chen2020uniter,li2020oscar,radford2021clip,kim2021vilt,li2021albef,wang2021vlmo} and fine-tune it on every single task independently.
Such an approach suffers from two limitations.
\textbf{\em (1) Low parameter efficiency}: 
Each real-world application requires the deployment of its dedicated fine-tuned model, where there is no parameter or inference computation sharing. 
This leads to a linearly increasing storage and inference compute redundancy in the long run.
{\bf\em (2) Lack of inter-task relatedness}: 
Though the fashion tasks are heterogeneous in nature, the fundamental components of the models are closely related in that all tasks require a deep content (image/sentence) understanding.
Exploiting the shared information across tasks thus
has the potential to improve model generalization capability leading to a performance boost.

Perhaps a natural solution would be applying Multi-Task Learning (MTL) \cite{crawshaw2020mtl_survey}.
However, most existing multi-task training methods~\cite{chen2018gradnorm,sener2018mgda,kendall2018uncertainty,liu2020imtl,navon2022nash-mtl} are designed for homogeneous tasks (\ie, one dataset with multi-task labels) and thus cannot be directly applied to the heterogeneous fashion tasks.
In our case, we are facing two challenges in building the fashion-domain MTL model: 
(1) \textit{Architecturally}, it is non-trivial to model the diverse tasks in one unified architecture.
Taking the popular CLIP \cite{radford2021clip} as an example, its two-stream architecture is designed for image-text alignment~\cite{ma2022eiclip} and thus lacks the modality fusion mechanism as required by many V+L fashion tasks (\eg, text-guided image retrieval~\cite{wu2021fashioniq,baldrati2022conditioned} and image captioning~\cite{yang2020facad}).
(2) 
In terms of \textit{optimization}, 
a fashion-domain MTL model is prone to the notorious \textit{negative transfer} problem~\cite{chen2018gradnorm,sener2018mgda,kendall2018uncertainty,liu2020imtl,navon2022nash-mtl,crawshaw2020mtl_survey} due to both task input/output format differences and imbalanced dataset sizes.
To the best of our knowledge, there has been no attempt at V+L MTL for the fashion domain. 

In this work,  we introduce a novel \textbf{FA}shion-focused \textbf{M}ulti-task \textbf{E}fficient learning method
for various \textbf{Vi}sion-and-\textbf{L}anguage based fashion tasks, dubbed as \textbf{\modelname}.
It achieves superior performance across a set of diverse fashion tasks with much fewer parameters as in Fig.~\ref{fig:trailer}.
Specifically, we design a task-versatile architecture on top of a pre-trained generic V+L model (\ie, CLIP~\cite{radford2021clip}).
To adapt the simple two-stream architecture of CLIP to various fashion tasks, we introduce a lightweight \textbf{\em Cross-Attention Adapter (XAA)} to enable the cross-modality interaction between the two streams. 
It makes the model flexible to support multiple task modes (\eg,
contrastive mode for retrieval, fusion mode for understanding, and generative mode for generation).
%
To address the negative transfer challenge,
we introduce a \textbf{\textit{Task-Specific Adapter (TSA)}} to absorb
inter-task input/output format incompatibilities by introducing
lightweight additional per-task parameters.
For further handling the dataset imbalance problem, 
a \textbf{\em multi-teacher distillation} scheme~\cite{clark2019bam} is formulated for our heterogeneous MTL problem.
It leverages the pre-trained per-task teachers to guide the optimization of our multi-task model, mitigating the overfitting risks of those tasks with smaller training dataset sizes.

Our {\bf \em contributions} are summarized as follows: 
\textbf{(I)} For the first time, we investigate the problem of multi-task learning on heterogeneous fashion tasks, eliminating the parameter redundancy and exploiting the inter-task relatedness. 
\textbf{(II)}
We propose \modelname with two novel adapters, adapting a pre-trained CLIP model to all tasks.
\textbf{(III)}
We introduce an efficient and effective multi-task training strategy supporting heterogeneous task modes in one unified model.
\textbf{(IV)}
Comprehensive experiments on four diverse fashion tasks (\ie, cross-modal retrieval~\cite{rostamzadeh2018fashiongen,ma2022eiclip},
text-guided image retrieval~\cite{vo2019tirg,wu2021fashioniq}, multi-modal classification~\cite{rostamzadeh2018fashiongen,zhuge2021kaleido}, and
image captioning~\cite{yang2020facad}) show that our method significantly outperforms the previous single-task state-of-the-art with 61.5\%
parameter saving (see Fig.~\ref{fig:trailer}).

\begin{figure}[t]
  \centering
\includegraphics[width=\linewidth]{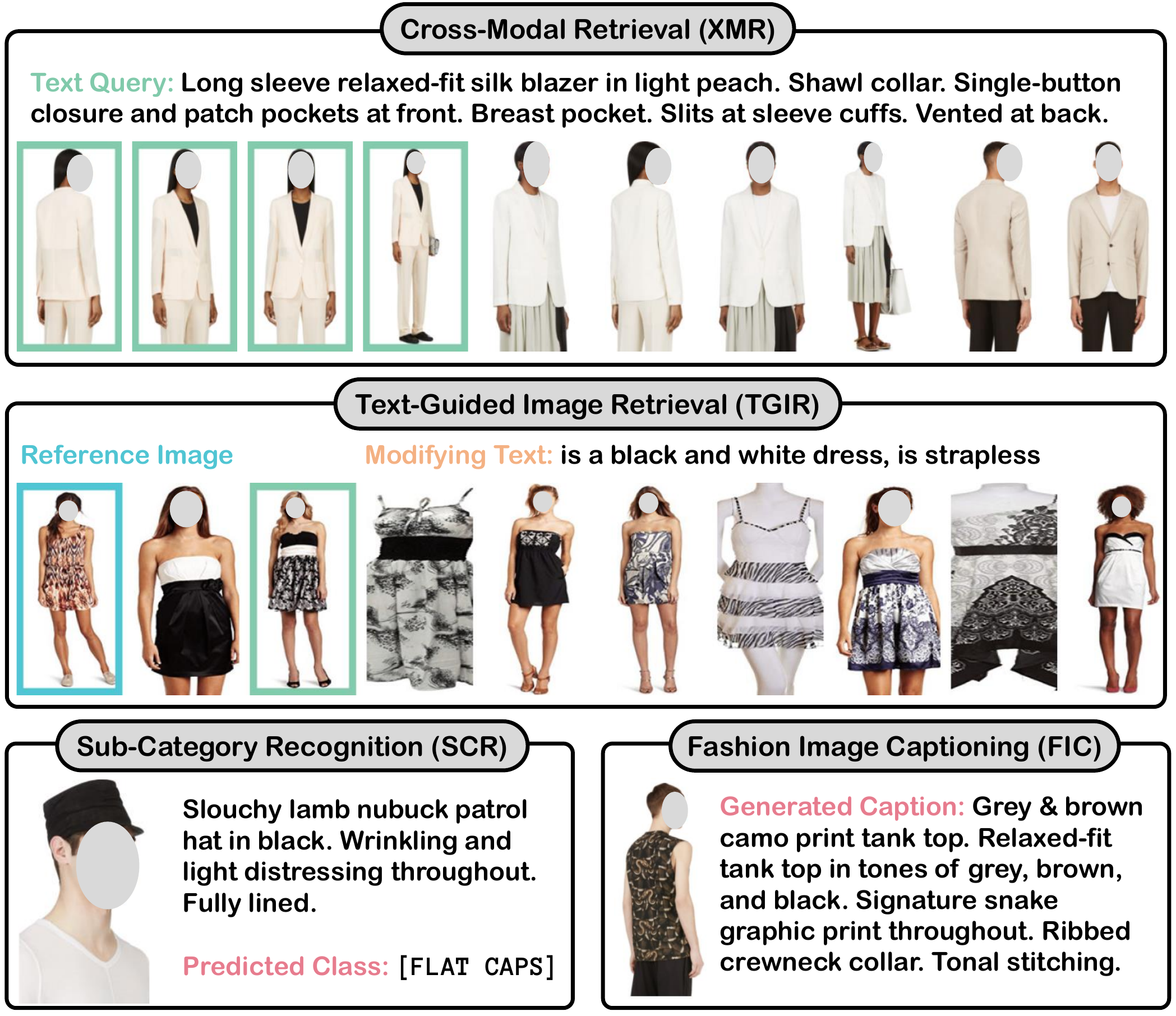}
   \caption{An illustration of four diverse fashion V+L Tasks studied in this work: cross-modal retrieval, text-guided image retrieval, sub-category recognition, and fashion image captioning. 
   Note, all predictions shown in this figure are made by our \modelname.
   \textcolor{ForestGreen}{Green box} indicates the ground truth matches of retrieval tasks.}
   \label{fig:tasks}
\end{figure}
\section{Related work}
\statement{Vision-Language Pre-training (VLP).}
With the advent of Transformers \cite{vaswani2017transformer,devlin2018bert,dosovitskiy2020vit},
many pioneer studies~\cite{li2019visualbert,lu2019vilbert,su2019vlbert,chen2020uniter,li2020oscar,kim2021vilt,li2021albef,huang2021soho,zhang2021vinvl} have demonstrated that VLP is effective in boosting various downstream V+L tasks in the generic domain. 
Since then, we have witnessed further developments of VLP methods, being bigger~\cite{radford2021clip,jia2021align,fei2022wenlan-nature}, more unified~\cite{wang2021vlmo,yu2022coca,singh2022flava,wang2022beitv3,wang2022ofa,lu2022unified-io} and more flexible~\cite{wang2021ufo,dou2022fiber,xu2022bridgetower}.

\statement{Fashion V+L tasks.} 
There exist a variety of heterogeneous tasks in the fashion domain.
As depicted in Fig.~\ref{fig:tasks}, we consider four popular fashion tasks in this work:
(1) \textit{Cross-Modal Retrieval (XMR)}
requests to efficiently retrieve the most matched image/sentence from a large candidate pool given a text/image query~\cite{rostamzadeh2018fashiongen,ma2022eiclip}.
(2) \textit{Text-Guided Image Retrieval (TGIR)} is a special type of image retrieval with a multi-modal query (a combination of a reference image and a modifying text) matched against a set of images~\cite{chen2020val,lee2021cosmo,kim2021dcnet,shin2021rtic,han2022uigr,delmas2021artemis}.
It not only requires a strong fusion of the reference image and modifying text, but also an efficient matching between the fused representation and all candidate images~\cite{wu2021fashioniq,han2022fashionvil}.
(3) \textit{Sub-Category Recognition (SCR)} requires an accurate class prediction made upon the fusion of an image-text pair~\cite{rostamzadeh2018fashiongen,zhuge2021kaleido}.
(4) \textit{Fashion Image Captioning (FIC)} generates a caption to describe the given image with semantically meaningful, fine-grained, and accurate words~\cite{yang2020facad}.
Many recent works have been trying to address these fashion tasks through VLP~\cite{gao2020fashionbert,zhuge2021kaleido,goenka2022fashionvlp,han2022fashionvil,yu2022commercemm,ji2022mvlt,mirchandani2022fad-vlp}.
Most of them focus on the pre-training, then simply fine-tune the pre-trained model on each downstream task independently. 
In contrast, we integrate all these tasks into a unified architecture and thus no separate fine-tuning is needed.
Since our fashion data is also abundant, most early works pre-train on the fashion domain directly. 
However, a number of recent works~\cite{baldrati2022conditioned,baldrati2022effective,dodds2022yahoo,chia2022fashionclip,ma2022eiclip} suggest that a generic-domain pre-trained CLIP~\cite{radford2021clip} generalizes even better on the fashion tasks. 
In this work, we also exploit a pre-trained CLIP model.
Different from the existing methods, we use a single multi-task learned model for all mentioned fashion tasks during fine-tuning.

\statement{Parameter-efficient tuning.}
Due to the increase in the size of V+L models, 
there is a growing interest in developing parameter-efficient methods to quickly adapt a large pre-trained model to specific tasks by using as few extra parameters as possible.
The most representative methods are adapters~\cite{houlsby2019nlp-adapter,chen2022adaptformer,sung2022vl-adapter,chen2022vitadapter},
prompt tuning~\cite{liu2021prompt-survey,zhou2022coop,zhou2022cocoop,jia2022vpt},
low-rank adaptation~\cite{hu2021lora} and their unified variants~\cite{he2022unified,mao2022unipelt,zhang2022noah}. 
Interestingly, whilst MTL can save much larger parameters, it is under-studied in V+L modeling. 
In this work, we propose two kinds of adapters (XAA and TSA in Sec.~\ref{sec:model_arch}) to adapt
CLIP specifically designed for MTL in the fashion domain.
Besides parameter-efficiently adapting CLIP,
our proposed adapters also serve as the key component for
task-versatile architecture design and enabling stable MTL.

\statement{Multi-task learning.}
Although MTL offers many advantages like improved data efficiency and reduced over-fitting, how to avoid negative transfer and cope with severely imbalanced dataset sizes is still an open question.
One common solution is to weight per-task losses or combine per-task gradients into a joint update direction using various heuristics~\cite{chen2018gradnorm,sener2018mgda,kendall2018uncertainty,liu2020imtl,navon2022nash-mtl}.
These works require the MTL model to have at least one forward propagation on each task so that they can manipulate the overall losses or gradients.
However, since V+L tasks are typically heterogeneous (especially in the fashion domain), this requirement cannot be easily satisfied, making these methods not directly applicable.
In contrast, Task Sampling-based MTL (TS-MTL) is without such a requirement and thus being widely adopted in V+L models~\cite{nguyen2019hdc,lu202012-in-1,hu2021unit,chen2020uniter,singh2022flava}. In TS-MTL, only one task along with its data point is sampled per iteration.
Since then, the heuristic task-sampling strategies~\cite{lu202012-in-1,hu2021unit,jean2019adaptive_scheduling} have been proposed, aiming to balance different tasks, avoiding the \textit{over-fitting} on easier (or data-poor) tasks or \textit{catastrophic forgetting}~\cite{french1999catastrophic} on harder (or data-rich) tasks. 
However, it is found that TS-MTL on its own often underperforms single-task trained models; it is thus typically followed by an additional per-task fine-tuning step~\cite{lu202012-in-1,hu2021unit}.
In this work, we formulate an effective knowledge distillation with multiple single-task teachers~\cite{clark2019bam} to alleviate the negative transfer without further fine-tuning on each task.

\begin{figure}[t]
\begin{center}
\includegraphics[width=0.9\linewidth]{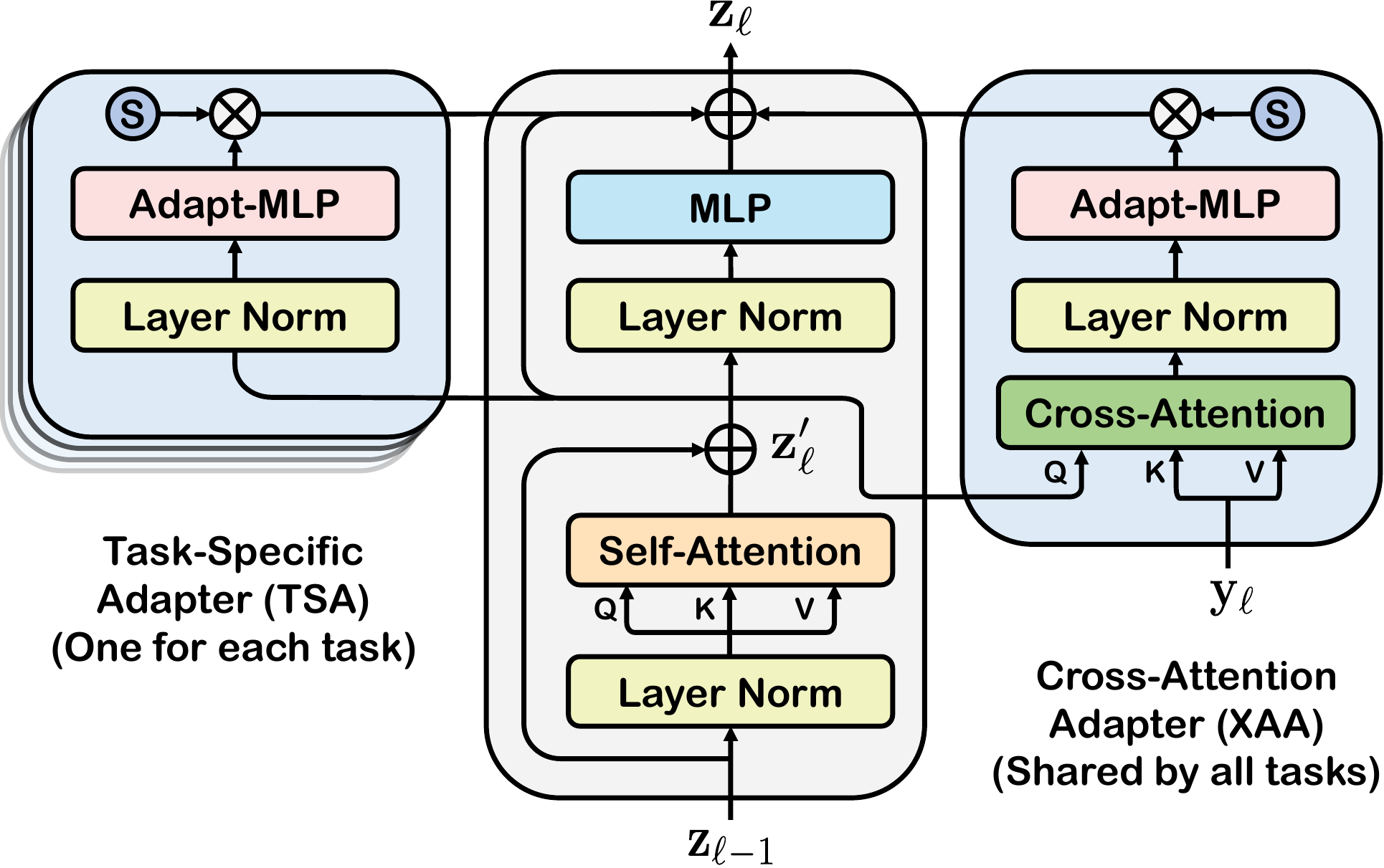}
\end{center}
\caption{An illustration of a task-versatile Transformer layer equipped with two newly-introduced adapters: cross-attention adapter (XAA) and task-specific adapter (TSA).}
\label{fig:adapter_arch}
\end{figure}
\section{Methodology}
We aim to address the most popular fashion tasks (shown in Fig.~\ref{fig:tasks}) using one single unified model. 
We introduce the details of our proposed \modelname as follows.

\subsection{Model architecture}
\label{sec:model_arch}
\modelname consists of a language encoder and a vision encoder initialized from the pre-trained CLIP (ViT-B/16 version)~\cite{radford2021clip}, as well as a set of newly introduced adapters.
\statement{Transformer layers.}
The key component in CLIP is the transformer backbone~\cite{vaswani2017transformer,dosovitskiy2020vit}.
A vanilla Transformer encoder consists of an input embedding layer (word embedding for language input and patch embedding for vision input) and several alternating layers made of Multi-Head Self-Attention (MHSA) and MLP blocks.
Layer Normalization (LN)~\cite{ba2016ln} is applied before every block, and residual connections after every block~\cite{dosovitskiy2020vit,radford2021clip}.
As shown in the middle of Fig.~\ref{fig:adapter_arch}, this process can be formulated as follows:
\begin{align}
\mathbf{z}_{0} &=\operatorname{Embedding}(\mathbf{x}), & \mathbf{z}_{0} & \in \mathbb{R}^{N \times D}, \\
\mathbf{z}_{\ell}^{\prime} &=\operatorname{MHSA}(\operatorname{LN}(\mathbf{z}_{\ell-1}))+\mathbf{z}_{\ell-1}, & \ell &=1 \ldots L, \\
\mathbf{z}_{\ell} &=\operatorname{MLP}(\operatorname{LN}(\mathbf{z}_{\ell}^{\prime}))+\mathbf{z}_{\ell}^{\prime}, & \ell &=1 \ldots L, \label{eaq:final_output}
\end{align}
where $N, D, L$ denotes the number of input tokens, transformer dimension, and the number of layers, respectively.

\statement{Proposed adapters.}
To adapt the original Transformer layers in CLIP to different fashion downstream tasks, we design two kinds of adapters in architecture design: 
\textbf{(1)} Task-Specific Adapter (TSA) 
for accommodating the non-shareable characteristics
of each individual task
(Fig.~\ref{fig:adapter_arch} left);
\textbf{(2)} Cross-Attention Adapter (XAA) for enabling the interaction between different modalities (Fig.~\ref{fig:adapter_arch} right).

For TSA we adopt the scaled parallel adapter
~\cite{he2022unified,chen2022adaptformer} that adds another bottleneck MLP (AdaptMLP) in parallel with the original MLP of each transformer layer.
Given an immediate input $\mathbf{z}_{\ell}^{\prime}$, it produces the adapted features, $\mathbf{z}_{\ell}^{tsa}$, via:
\begin{equation}
\mathbf{z}_{\ell}^{tsa} = s \cdot \operatorname{AdaptMLP}(\operatorname{LN}(\mathbf{z}_{\ell}^{\prime})),
\end{equation}
where $s$ represents a learnable scale.

We construct an XAA module by further adding another Multi-Head Cross Attention (MHXA) layer~\cite{lu2019vilbert,tan2019lxmert,dou2022fiber} at the bottom of a TSA.
Specifically, this MHXA uses the hidden state of the current stream $\mathbf{z}_{\ell}^{\prime}$ as the queries and the output 
$\mathbf{y}_{\ell}$ (\eg, hidden state after MHSA or MLP) of another stream as the keys and values.
The attended cross-modality features $\mathbf{z}_{\ell}^{xaa}$ are computed via:
\begin{equation}
\mathbf{z}_{\ell}^{xaa} = s \cdot \operatorname{AdaptMLP}(\operatorname{LN}(\operatorname{MHXA}(\mathbf{z}_{\ell}^{\prime}, \mathbf{y}_{\ell}))).
\end{equation}
Within this mechanism, our XAA can exchange the information between different modalities.

Finally, both $\mathbf{z}_{\ell}^{tsa}$ and $\mathbf{z}_{\ell}^{xaa}$ are added up to the original output via residual connections.
We thus rewrite Eq.~\eqref{eaq:final_output} as:
\begin{equation}
\mathbf{z}_{\ell} =\operatorname{MLP}(\operatorname{LN}(\mathbf{z}_{\ell}^{\prime}))+\mathbf{z}_{\ell}^{\prime} + \mathbf{z}_{\ell}^{tsa} + \epsilon \cdot \mathbf{z}_{\ell}^{xaa}, \epsilon \in \left\{0, 1\right\},
\end{equation}
where $\epsilon$ represents a gate that can turn on/off XAA according to the task need.

\begin{figure*}[t]
\begin{center}
\includegraphics[width=0.95\linewidth]{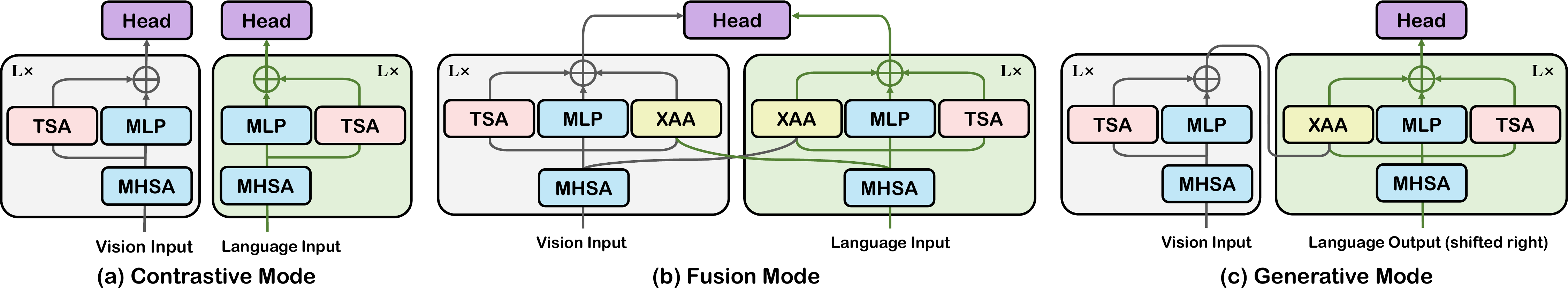}
\end{center}
\caption{
Schematic overview of three operational modes with our \modelname.
XAA: Cross-Attention Adapter;
TSA: Task-Specific Adapter.
Layer normalization and original residual connections are not shown here for simplicity.
}
\label{fig:whole_pipeline}
\end{figure*}

\statement{Operational modes and fashion tasks.}
Our \modelname can
switch among three operational modes to flexibly support various fashion tasks (see Fig.~\ref{fig:whole_pipeline}).

$\diamond$~\textbf{Contrastive mode}:
This mode supports {\em Cross-Modal Retrieval} (XMR) tasks, including both text-to-image and image-to-text retrieval~\cite{ma2022eiclip,han2022fashionvil}.
All XAA modules are turned off as no need for inter-modal interaction, whereas only TSA modules are applied as in Fig.~\ref{fig:whole_pipeline}(a).
During training, given a batch of $B$ image-text pairs $(\mathbf{I}$, $\mathbf{T})$ = $\{(I_1, T_1),\dots,(I_B, T_B)\}$, we first compute their unimodal representations by the TSA-equipped vision and language encoders independently. 
Then, we maximize their similarities via symmetrical contrastive loss:
\begin{equation}
    \mathcal{L}_{\mathrm{XMR}} = \frac{1}{2} \left[\mathcal{L}_{\mathrm{InfoNCE}}(\mathbf{T}, \mathbf{I}) + \mathcal{L}_{\mathrm{InfoNCE}}(\mathbf{I}, \mathbf{T})\right],
\end{equation}
\begin{equation}
\footnotesize
\mathcal{L}_{\mathrm{InfoNCE}}(\mathbf{X}, \mathbf{Y})=- \frac{1}{B} \sum_{i=1}^{B} \log \frac{\exp (s(X_i, Y_i) / \tau)}{\sum_{j=1}^{B} \exp (s(X_i, Y_j) / \tau)},
\label{eqa:infonce}
\end{equation}
where $\tau$ is a learnable temperature.
The similarity is measured by the dot product of their pooled then normalized features: $
s(I_{i}, T_{j})=f_{\theta}^{[c]}(I_{i})^{T} \cdot f_{\theta}^{[c]}(T_{j}).
$

$\diamond$~\textbf{Fusion mode}:
As in Fig.~\ref{fig:whole_pipeline}(b), both XAA and TSA modules are enabled in this mode.
Given an input image-text pair $(I, T)$,
the model serves as a single-stream fusion encoder producing two cross-modal attended representations: $f_{\theta}^{[f]}([I;T])$ and $ f_{\theta}^{[f]}([T;I])$.
The final fused representation is a simple addition: $f_{\theta}^{[f]}(I, T) = f_{\theta}^{[f]}([I;T]) + f_{\theta}^{[f]}([T;I])$~\footnote{More complex fusion methods (\eg, \cite{baldrati2022conditioned}) may yield better results. We leave this for future study.}.
This mode is useful for the {\em Sub-Category Recognition} (SCR)~\cite{rostamzadeh2018fashiongen,zhuge2021kaleido} and {\em Text-Guided Image Retrieval} (TGIR)~\cite{vo2019tirg,wu2021fashioniq}.

SCR aims to predict the subcategory of fashion products based on both input image and text.
We thus append a classifier on top of the fused representation and minimize its cross-entropy loss:
\begin{equation}
    \mathcal{L}_{\mathrm{SCR}}=-\mathbb{E}_{(I, T) \sim D} \log P\left(f_{\theta}^{[f]}(I, T)\right).
\end{equation}

Considering the unique challenges of TGIR (\ie, requiring not only strong fusion but also efficient matching), \modelname operates in a hybrid mode for it.
Formally, given a batch of \{reference images $\mathbf{I}^r$, modifying text $\mathbf{T}$, target images $\mathbf{I}^t$\}, we first calculate the fused representation $f_{\theta}^{[f]}(I^r, T)$ in the fusion mode; Then, we obtain the target image representation $f_{\theta}^{[c]}(I^t)$ in the contrastive mode.
During training, we pull them closer in a contrastive way:
\begin{equation}
    \mathcal{L}_{\mathrm{TGIR}} = \mathcal{L}_{\mathrm{InfoNCE}}\left((\mathbf{I}^r, \mathbf{T}), \mathbf{I}^t\right),
\end{equation}
where the similarity is still measured by dot product: $s((I_i^r, T_i), I_j^t) = f_{\theta}^{[f]}(I_{i}^r, T_i)^{T} \cdot f_{\theta}^{[c]}(I_{j}^t)$.

$\diamond$~\textbf{Generative mode}:
This works as a \texttt{seq2seq} model 
performing the generative tasks auto-regressively~\cite{sutskever2014seq2seq,cho2014rnnseq2seq}, \eg, {\em Fashion Image Captioning} (FIC)~\cite{yang2020facad}.
As in Fig.~\ref{fig:whole_pipeline}(c), we use a TSA-equipped vision encoder as the encoder, a TSA-equipped language encoder as the decoder, and only image-to-text XAA modules for the conditional caption synthesis.
Following a standard encoder-decoder architecture, our image encoder provides \textit{layer-wise} latent memory and the text decoder learns to maximize the conditional likelihood of the paired text under the forward auto-regressive factorization~\cite{yu2022coca,li2022blip}:
\begin{equation}
    \mathcal{L}_{\mathrm{FIC}}=
    -\mathbb{E}_{(I, T) \sim D} 
    \sum_{a=1}^{A}
    \log P\left(T_a\middle|f_{\theta}^{[g]}(I;T_{<a})\right),
\end{equation}
where $A$ denotes the length of each sentence, and $f_{\theta}^{[g]}([I;T])$ denotes the representation.
During training, we enforce the teacher forcing~\cite{williams1989teacherforcing} to achieve parallel computation and thus maximize the learning efficiency.

\subsection{Heterogeneous multi-task learning}
\label{sec:mtl}
For training $\mathcal{T}$ fashion tasks, we need to optimize both task-agnostic parameters $\theta_s$ (\ie, CLIP backbone \& XAA modules) and a set of task-specific components $\left\{\theta_t\right\}_{t=1}^{\mathcal{T}}$ (\ie, TSA modules \& task heads).
Our objective is to maximize the overall performance across all tasks.
The heterogeneity nature of fashion tasks
causes the discrepancy in mini-batch construction and training dynamics (\eg, converging speed, overfitting) as well as data imbalance,
making our multi-task learning particularly challenging.
To address all these issues, we exploit the idea of Multi-Teacher Distillation (MTD)~\cite{hinton2015kd,clark2019bam}.

Specifically, MTD consists of two stages.
In the {\bf\em first stage}, we train a teacher model with the identical architecture as our multi-task model on every task.
Then in the {\bf\em second stage}, we apply these teachers 
to guide the training of the multi-task model (\ie, the student) with designed per-task distillation objectives.

For XMR, we first compute the image-text similarity using the features of the single-task teacher $g_{\mathrm{xmr}}$:
$
\Tilde{s}(I_{i}, T_{j})=g_{\mathrm{xmr}}(I_{i})^{T} \cdot g_{\mathrm{xmr}}(T_{j}).
$
Using this similarity as a pseudo-target, we maximize its mutual information with the student's counterpart~\cite{li2021albef}:
\begin{equation}
\small
\mathcal{L}_{\mathrm{XMR}}^{\mathrm{D}} = 
\frac{1}{2B}
\sum_b^B \left(\operatorname{KL}\left(\mathbf{s}_{b, \cdot}\parallel \Tilde{\mathbf{s}}_{b, \cdot}\right)
+
\operatorname{KL}\left(\mathbf{s}_{\cdot, b}\parallel \Tilde{\mathbf{s}}_{\cdot, b}\right)\right),
\end{equation}
where $\operatorname{KL}(\cdot\parallel\cdot)$ denotes the Kullback–Leibler divergence loss on the softmax of the inputs.

For TGIR, we use a similar method as XMR to distill the knowledge from single-task teacher $g_{\mathrm{tgir}}$:
\begin{equation}
\mathcal{L}_{\mathrm{TGIR}}^{\mathrm{D}} = 
\frac{1}{B}
\sum_b^B \operatorname{KL}\left(\mathbf{s}_{(b,b), \cdot}\parallel \Tilde{\mathbf{s}}_{(b,b), \cdot}\right),
\end{equation}
where the soft target is calculated via: 
$\Tilde{s}((I_i^r, T_i), I_j^t) = g_{\mathrm{tgir}}(I_{i}^r, T_i)^{T} \cdot g_{\mathrm{tgir}}(I_{j}^t)$.

For SCR and FIC, we directly use the classification probabilities predicted by the teachers as pseudo-targets:
\begin{equation}
\mathcal{L}_{\mathrm{SCR}}^{\mathrm{D}} = 
\operatorname{KL}\left(
f_{\theta}^{[f]}(I, T)
\parallel
g_{\mathrm{scr}}(I, T)
\right),
\end{equation}
\begin{equation}
\mathcal{L}_{\mathrm{FIC}}^{\mathrm{D}} = 
\sum_{a=1}^{A}
\operatorname{KL}\left(
f_{\theta}^{[g]}(I;T_{<a})_a
\parallel
g_{\mathrm{fic}}(I;T_{<a})_a
\right).
\end{equation}

\statement{Task scheduling.}
For training simplicity, we randomly sample one task per iteration. 
We optimize the summation of the original loss and distillation loss as:  
\begin{equation}
\mathcal{L} = \mathcal{L}_{\mathrm{[task]}} + \mathcal{L}_{\mathrm{[task]}}^{\mathrm{D}},
\quad
\scriptstyle
\mathrm{[task]} \overset P\sim \{\mathrm{XMR, TGIR, SCR, FIC}\},
\end{equation}
where $P$ denotes the sampling probability.
To tackle data imbalance, unless stated otherwise we set the sampling probability for a particular task $\tau$ linearly proportional to the size of its dataset $|D_{\tau}|$~\cite{sanh2019hierarchical,hu2021unit}.
We name this strategy as {\bf \em size-proportional sampling}.

\section{Experiments}
\statement{Datasets.}
We evaluate our model on the datasets commonly used in the previous methods.
Specifically, we test FashionGen~\cite{rostamzadeh2018fashiongen} for XMR, SCR, and FIC, and FashionIQ~\cite{wu2021fashioniq} for TGIR.
FashionGen~\cite{rostamzadeh2018fashiongen} contains $68k$ fashion products accompanied by text descriptions. 
Each product includes $1\sim6$ images from different angles, resulting in $260.5k$ image-text pairs for training, and $35.5k$ for testing.
FashionIQ contains $18k$ training triplets (\ie, reference image, modifying text, target image) and $6k$ validation triplets over three categories: Dress, Shirt, and Toptee. 
Each pair (reference image, target image) is manually annotated with two modifying texts, which are concatenated~\cite{wu2021fashioniq}.

\statement{Implementation details.}
We use MMF~\cite{singh2020mmf} and PyTorch \cite{paszke2019pytorch} to implement our \modelname.
We use the off-the-shelf CLIP from HuggingFace~\cite{wolf2020huggingface} as our pre-trained model.
We use 4 RTX 3090 GPUs for the multi-task training.
The default bottleneck dimension of AdaptMLP is $64$.
More implementation details are listed in the supplementary file.

\statement{Performance metrics.}
Following~\cite{zhuge2021kaleido,han2022fashionvil}, we report R@K for retrieval, Accuracy \& Macro$F_1$ for classification and BLEU-4~\cite{papineni2002bleu} \& METEOR~\cite{banerjee2005meteor} \& ROUGE-L~\cite{lin2004rouge} \& CIDEr~\cite{vedantam2015cider} for captioning.
{\bf (1)} For each task, we first report {\bf \em the average absolute performance}: $\mu_{\mathcal{T}_{i}} = \frac{1}{|M|} \sum_{j=0}^{|M|} {M_{\mathcal{T}_{i}, j}}$.
{\bf (2)} Since there is no unified objective among multiple tasks and the scale of per-task metrics often varies largely, we then report 
{\bf \em the average per-task relative performance} $\Delta_{\mathcal{T}_{i}}$ \wrt the single-task baseline:
$
\Delta_{\mathcal{T}_{i}}=\left(\mu_{\mathcal{T}_{i}}-\mu_{\mathrm{STL}}\right) / \mu_{\mathrm{STL}}.
$
This can clearly indicate the positive/negative transfer effect.
{\bf (3)} We also report {\bf \em the relative parameters saving} of \modelname and its variants \wrt
the vanilla CLIP baseline.
Note, inference speed comparison is infeasible 
as this depends on the application of different tasks (no fixed rules).
 
\begin{table}[t]
\centering
\resizebox{0.98\columnwidth}{!}{%
\begin{tabular}{l?ccccccc}
\toprule[1.5pt]
\multirow{2}{*}{\textbf{Methods}} & \multicolumn{3}{c}{\textbf{Image to Text}}       & \multicolumn{3}{c}{\textbf{Text to Image}}       & \multirow{2}{*}{\textbf{Mean}} \\
                                  & \textbf{R@1}   & \textbf{R@5}   & \textbf{R@10}  & \textbf{R@1}   & \textbf{R@5}   & \textbf{R@10}  &                                \\ \midrule[1.5pt]
\textbf{FashionBERT}~\cite{gao2020fashionbert}              & 23.96          & 46.31          & 52.12          & 26.75          & 46.48          & 55.74          & 41.89                          \\
\textbf{OSCAR}~\cite{li2020oscar}                    & 23.39          & 44.67          & 52.55          & 25.10          & 49.14          & 56.68          & 41.92                          \\
\textbf{KaledioBERT}~\cite{zhuge2021kaleido}              & 27.99          & 60.09          & 68.37          & 33.88          & 60.60          & 68.59          & 53.25                          \\
\textbf{EI-CLIP}~\cite{ma2022eiclip}              & 38.70          & 72.20          & 84.25          & 40.06          & 71.99          & 82.90          & 65.02                          \\
\textbf{MVLT}~\cite{ji2022mvlt}              & 33.10          & 77.20          & 91.10          & 34.60         & 78.00          & 89.50          & 67.25                          \\
\textbf{FashionViL}~\cite{han2022fashionvil}               & 65.54          & 91.34          & 96.30          & 61.88          & 87.32          & 93.22          & 82.60                          \\
\textbf{FashionViL}(\textit{vit})          & 63.74         &  90.02           & 95.98         &  60.76        &  86.18       &  92.96        &  81.61                       \\ \midrule[0.5pt]
\textbf{\modelname}({\em ST})                 & 65.02          & 90.96          & 96.20          & \textbf{63.56}          & 86.84          & 93.06          & 82.61                          \\
\textbf{\modelname}               & \textbf{65.94} & \textbf{91.92} & \textbf{97.22} & 62.86          & \textbf{87.38}          & \textbf{93.52}          & \textbf{83.14}                 \\ \bottomrule[1.5pt]
\end{tabular}%
}
\caption{Cross-Modal Retrieval (XMR) results on FashionGen~\cite{rostamzadeh2018fashiongen}. 
Test protocol: \texttt{random 100} \cite{gao2020fashionbert,zhuge2021kaleido,han2022fashionvil}.
}
\label{tab:xmr_results}
\end{table}
\subsection{Comparisons with prior art methods}
\label{sec:SOTA}
We compare our models with the previous state-of-the-art methods on each task. 
For extensive and fair comparisons, all the prior competitors are based on large-scale pre-trained models.
We even implement an enhanced variant of the latest art model FashionViL~\cite{han2022fashionvil} by replacing ResNet50~\cite{he2016resnet} with the ViT-B/16~\cite{dosovitskiy2020vit} backbone (same as our \modelname), denoted as \texttt{FashionViL(\textit{vit})}.
In design, all the previous methods adopt Single-Task Learning (STL). 
We compare them with two variants of our model:
{\bf (1)} single unified MTL model;
{\bf (2)} STL variant of our \modelname, denoted as \texttt{FAME-ViL(\textit{ST})}, which is trained on each task independently using the same TSA and XAA design as \modelname.

\statement{XMR evaluation.}
We consider both image-to-text retrieval and text-to-image retrieval with two kinds of protocols used by previous methods:
\textbf{(1)} \texttt{random 100}: 
For each query, 100 candidates are randomly sampled from the same category to construct a retrieval database; The goal is to locate the positive match depicting the same garment instance from these 100 same-category negative matches\cite{gao2020fashionbert,zhuge2021kaleido,ji2022mvlt}.
\textbf{(2)} \texttt{full database}:
We also adopt a more challenging and practical protocol that conducts retrieval on the entire product set \cite{ma2022eiclip,han2022fashionvil}, which is in line with actual product retrieval scenarios.
We use \texttt{random 100} to compare with prior art methods while using \texttt{full database} to do ablation studies.
The results of XMR on FashionGen \cite{rostamzadeh2018fashiongen} are reported in Tab.~\ref{tab:xmr_results}.
We draw several observations:
\textbf{(1)} Our \modelname outperforms all prior art fashion models often by a large margin, validating the performance advantages of our method over alternatives in addition to better parameter efficiency.
\textbf{(2)} \modelname is superior over its single-task variant FAME-ViL({\em ST}) in most cases and on the average accuracy, suggesting that our multi-task learning strategy is effective in exploiting the inter-task relatedness.
\textbf{(3)} Our FAME-ViL({\em ST}) can surpass all prior models
pre-trained on large fashion domain data~\cite{gao2020fashionbert,li2020oscar,zhuge2021kaleido, han2022fashionvil}, suggesting that using fashion data in pre-training is not necessarily most important, and model design (\eg, our TSA and XAA) could play a more significant role. 
Similarly, its large margin over the previous pre-trained CLIP-based model~\cite{ma2022eiclip} further validates the significance of model architecture design. 

\begin{table}[t]
\centering
\resizebox{\columnwidth}{!}{%
\begin{tabular}{l?ccccccc}
\toprule[1.5pt]
\multirow{2}{*}{\textbf{Methods}} & \multicolumn{2}{c}{\textbf{Dress}}                                      & \multicolumn{2}{c}{\textbf{Shirt}}                                      & \multicolumn{2}{c}{\textbf{Toptee}}                                     & \multirow{2}{*}{\textbf{Mean}}     \\
                                  & \textbf{R@10}                      & \textbf{R@50}                      & \textbf{R@10}                      & \textbf{R@50}                      & \textbf{R@10}                      & \textbf{R@50}                      &                                    \\ \midrule[1.5pt]
\textbf{CIRPLANT}~\cite{liu2021cirr}                 & 17.45                              & 40.41                              & 17.53                              & 38.81                              & 21.64                              & 45.38                              & 30.20                              \\
\textbf{TIRG}(\textit{bert})~\cite{vo2019tirg}$\dagger$               & 27.17                              & 53.25                              & 22.28                              & 45.58                              & 27.84                              & 57.11                              & 38.87                              \\
\textbf{FashionVLP}~\cite{goenka2022fashionvlp}               & 26.77                              & 53.20                              & 22.67                              & 46.22                              & 28.51                              & 57.47                              & 39.14                              \\
\textbf{FashionViL}~\cite{han2022fashionvil}               & 33.47                              & 59.94                              & 25.17                              & 50.39                              & 34.98                              & 60.79                              & 44.12                              \\
\textbf{FashionViL}(\textit{vit})          & 31.53                                   & 57.91                                   & 26.74                                   & 50.69                                   & 36.77                                   & 61.81                                   & 44.24                                   \\
\textbf{Baldrati \etal} \cite{baldrati2022conditioned}                   & 33.81                              & 59.40                              & 39.99                              & 60.45                              & 41.41                              & 65.37                              & 50.07                              \\
\textbf{Zhao \etal} \cite{zhao2022progressive}                   & 33.60                              & 58.90                              & 39.45                              & 61.78                              & 43.96                              & 68.33                              & 51.00                              \\ \midrule[0.5pt]
\textbf{\modelname}({\em ST})                 & 37.78                            & 63.86                              & 45.63                              & 66.78                            & 47.22                             &  70.88                            &    55.36                           \\
\textbf{\modelname}             & \multicolumn{1}{c}{\textbf{42.19}} & \multicolumn{1}{c}{\textbf{67.38}} & \multicolumn{1}{c}{\textbf{47.64}} & \multicolumn{1}{c}{\textbf{68.79}} & \multicolumn{1}{c}{\textbf{50.69}} & \multicolumn{1}{c}{\textbf{73.07}} & \multicolumn{1}{c}{\textbf{58.29}} \\ \bottomrule[1.5pt]
\end{tabular}%
}
\caption{Text-Guided Image Retrieval (TGIR) results on FashionIQ~\cite{wu2021fashioniq}. 
$\dagger$: The results taken from \cite{han2022fashionvil}.
}
\label{tab:tgir_results}
\end{table}
\statement{TGIR evaluation.}
We compare our \modelname with TGIR-specialist methods~\cite{liu2021cirr,vo2019tirg,goenka2022fashionvlp,baldrati2022conditioned,zhao2022progressive} and the art fashion-focused V+L model FashionViL~\cite{han2022fashionvil} under the original protocol used by FashionIQ~\cite{wu2021fashioniq}. 
The results are given in Tab.~\ref{tab:tgir_results}.
We have similar observations as on XMR. 
In particular, we note that our single-task variant
already achieve a new art performance.
With a simple addition-based fusion mechanism, 
\modelname can even outperform significantly \cite{baldrati2022conditioned}
with the same CLIP pre-training and a complex fusion module.
We attribute this mostly to the contribution of XAA-backed inter-modal interaction (See Tab.~\ref{tab:ablation_study}).

\begin{table}[t]
\centering
\setlength{\tabcolsep}{5pt}
\resizebox{\columnwidth}{!}{%
\begin{tabular}{l?ccc?ccccc}
\toprule[1.5pt]
\multirow{2}{*}{\textbf{Methods}} & \multicolumn{3}{c?}{\textbf{SCR}}                & \multicolumn{5}{c}{\textbf{FIC}}                                                   \\
                                  & \textbf{Acc}   & $\textbf{F}_1$     & \textbf{Mean}  & \textbf{B}     & \textbf{M}     & \textbf{R}     & \textbf{C}     & \textbf{Mean}  \\ \midrule[1.5pt]
\textbf{FashionBERT}~\cite{gao2020fashionbert}$\dagger$    & 85.27          & 62.00          & 73.64          & 3.30           & 9.80           & 29.70          & 30.10          & 18.23          \\
\textbf{OSCAR}~\cite{li2020oscar}$\dagger$          & 84.23          & 59.10          & 71.67          & 4.50           & 10.90          & 30.10          & 30.70          & 19.05          \\
\textbf{KaleidoBERT}~\cite{zhuge2021kaleido}    & 88.07          & 63.60          & 75.84          & 5.70           & 12.80          & 32.90          & 32.60          & 21.00          \\
\textbf{FashionViL}~\cite{han2022fashionvil}     & 92.23          & 83.02          & 87.63          & 16.71         & \textbf{25.97}          & 37.82               &  39.08              & 29.90             \\
\textbf{MVLT}~\cite{ji2022mvlt}    & 93.57          & 82.90          & 88.24          & -           & -          & -          & -          & -          \\
\textbf{FashionViL}(\textit{vit}) & 94.01          & 85.77          & 89.89          & 16.18            & 25.60           & 37.23               &  39.30              & 29.58     \\ \midrule[0.5pt]
\textbf{\modelname}(\textit{ST})       & 94.33          & 86.21          & 90.27          & 29.97          & 24.83          & 54.79          & 145.1          & 63.67          \\
\textbf{\modelname}     & \textbf{94.67} & \textbf{88.21} & \textbf{91.44} & \textbf{30.73} & 25.04 & \textbf{55.83} & \textbf{150.4} & \textbf{65.50} \\ \bottomrule[1.5pt]
\end{tabular}%
}
\caption{Results of Subcategory Recognition (SCR) and Fashion Image Captioning (FIC) on FashionGen~\cite{rostamzadeh2018fashiongen}.
$\dagger$: copied from \cite{zhuge2021kaleido}.}
\label{tab:scr_cap_results}
\end{table}
\begin{table*}[ht]
\centering
\resizebox{\linewidth}{!}{%
\begin{tabular}{c?cl?ccccccccccc}
\toprule[1.5pt]
                                  & \multicolumn{2}{c?}{}                                   &                                          & \multicolumn{2}{c}{\textbf{$\mathcal{T}_1$: XMR}}                                    & \multicolumn{2}{c}{\textbf{$\mathcal{T}_2$: TGIR}}                                    & \multicolumn{2}{c}{\textbf{$\mathcal{T}_3$: SCR}}                                             & \multicolumn{2}{c}{\textbf{$\mathcal{T}_4$: FIC}}                                             &                                  &                                        \\
\multirow{-2}{*}{\textbf{Groups}} & \multicolumn{2}{c?}{\multirow{-2}{*}{\textbf{Methods}}} & \multirow{-2}{*}{\textbf{\#Params (\%)}} & $\mu$                         & $\Delta$                                  & $\mu$                         & $\Delta$                                   & $\mu$                                  & $\Delta$                                  & $\mu$                                  & $\Delta$                                  & \multirow{-2}{*}{\Large{$\bar{\mu}$}} & \multirow{-2}{*}{\Large{$\bar{\Delta}$}}       \\ \midrule[1.5pt]
                                  & (1)        & \textbf{STL}                               & \cellcolor[HTML]{FFFFC7}\textbf{0.0}     & \cellcolor[HTML]{FFFFC7}66.30 & \cellcolor[HTML]{FFFFC7}0.0            & \cellcolor[HTML]{FFFFC7}51.87 & \cellcolor[HTML]{FFFFC7}0.0             & \cellcolor[HTML]{FFFFC7}\textbf{90.34} & \cellcolor[HTML]{FFFFC7}\textbf{0.0}   & -                                      & -                                      & 52.13                            & 0.0                                    \\
                                  & (2)        & \textbf{STL + TSA}                         & \cellcolor[HTML]{FFE2E2}+1.35            & \textbf{69.99}                & \cellcolor[HTML]{C5FFC5}\textbf{+5.56} & 52.59                         & \cellcolor[HTML]{F6FFF6}+1.39           & 90.10                                  & \cellcolor[HTML]{FFDCDC}-0.27          & -                                      & -                                      & 53.25                            & \cellcolor[HTML]{F6FFF6}+1.67          \\
                                  & (3)        & \textbf{STL + XAA}                         & \cellcolor[HTML]{FFDCDC}+14.70           & 66.30                         & 0.0                                    & 53.83                         & \cellcolor[HTML]{F6FFF6}+3.78           & 89.89                                  & \cellcolor[HTML]{FFDCDC}-0.50          & \cellcolor[HTML]{FFFFC7}\textbf{63.70} & \cellcolor[HTML]{FFFFC7}\textbf{0.0}   & 68.43                            & \cellcolor[HTML]{F6FFF6}+0.82          \\
\multirow{-4}{*}{\makecell{\rom{1}\\(Sec.~\ref{sec:ablation_study})}}             & (4)        & \textbf{STL + TSA + XAA (FAME-ViL}(\textit{ST}))                   & \cellcolor[HTML]{FF9696}+15.96           & \textbf{69.99}                & \cellcolor[HTML]{C5FFC5}\textbf{+5.56} & \textbf{55.47}                & \cellcolor[HTML]{DAFFDA}\textbf{+6.94}  & 90.27                                  & \cellcolor[HTML]{FFDCDC}-0.07          & 63.67                                  & -0.05                                  & \textbf{69.85}                   & \cellcolor[HTML]{DAFFDA}\textbf{+3.10} \\ \midrule
                                  & (5)        & \textbf{MTL}                               & \cellcolor[HTML]{C5FFC5}\textbf{-70.43}  & 57.65                         & \cellcolor[HTML]{FF9696}-13.05         & 49.57                         & \cellcolor[HTML]{FF9696}-4.43           & 85.95                                  & \cellcolor[HTML]{FF9696}-4.86          & -                                      & -                                      & 48.29                            & \cellcolor[HTML]{FF9696}-5.59          \\
                                  & (6)        & \textbf{MTL + TSA}                         & \cellcolor[HTML]{C5FFC5}-70.11           & 67.97                         & \cellcolor[HTML]{F6FFF6}+2.52          & 52.04                         & \cellcolor[HTML]{F6FFF6}+0.33           & 90.32                                  & \cellcolor[HTML]{FFDCDC}-0.02          & -                                      & -                                      & 52.58                            & \cellcolor[HTML]{F6FFF6}+0.71          \\
                                  & (7)        & \textbf{MTL + XAA}                         & \cellcolor[HTML]{DAFFDA}-67.65           & 65.87                         & \cellcolor[HTML]{FFE2E2}-0.65          & 52.59                         & \cellcolor[HTML]{F6FFF6}+1.39           & \textbf{90.93}                         & \cellcolor[HTML]{F6FFF6}\textbf{+0.65} & 60.99                                  & \cellcolor[HTML]{FF9696}-4.25          & 67.60                            & \cellcolor[HTML]{FFDCDC}-0.72          \\
\multirow{-4}{*}{\makecell{\rom{2}\\(Sec.~\ref{sec:ablation_study})}}             & (8)        & \textbf{MTL + TSA + XAA (base MTL)}        & \cellcolor[HTML]{DAFFDA}-67.33           & \textbf{69.31}                & \cellcolor[HTML]{DAFFDA}\textbf{+4.54} & \textbf{55.41}                & \cellcolor[HTML]{F6FFF6}\textbf{+6.82}  & 90.84                                  & \cellcolor[HTML]{F6FFF6}+0.55          & \textbf{65.17}                         & \cellcolor[HTML]{F6FFF6}\textbf{+2.31} & \textbf{70.18}                   & \cellcolor[HTML]{DAFFDA}\textbf{+3.56} \\ \midrule
                                  & (9)        & \textbf{base MTL + MTD (\modelname)}             & \cellcolor[HTML]{DAFFDA}-67.33           & 70.00                         & \cellcolor[HTML]{C5FFC5}+5.56          & \textbf{58.29}                & \cellcolor[HTML]{C5FFC5}\textbf{+12.38} & \textbf{91.44}                         & \cellcolor[HTML]{C5FFC5}\textbf{+1.22} & 65.50                                  & \cellcolor[HTML]{DAFFDA}+2.83          & \textbf{71.31}                   & \cellcolor[HTML]{C5FFC5}\textbf{+5.50} \\
                                  & (10)       & \textbf{base MTL + MTD + Uniform}           & \cellcolor[HTML]{DAFFDA}-67.33           & 67.70                         & \cellcolor[HTML]{F6FFF6}+2.11          & 57.31                         & \cellcolor[HTML]{C5FFC5}+10.49          & 91.36                                  & \cellcolor[HTML]{C5FFC5}+1.13          & 65.12                                  & \cellcolor[HTML]{F6FFF6}+2.23          & 70.37                            & \cellcolor[HTML]{DAFFDA}+3.99          \\
                                  & (11)       & \textbf{base MTL + MTD + Round-robin}      & \cellcolor[HTML]{DAFFDA}-67.33           & 67.79                         & \cellcolor[HTML]{F6FFF6}+2.25          & 57.47                         & \cellcolor[HTML]{C5FFC5}+10.80          & 91.35                                  & \cellcolor[HTML]{C5FFC5}+1.12          & 64.87                                  & \cellcolor[HTML]{F6FFF6}+1.84          & 70.37                            & \cellcolor[HTML]{DAFFDA}+4.00          \\
                                  & (12)       & \textbf{base MTL + IAS~\cite{jean2019adaptive_scheduling}}               & \cellcolor[HTML]{DAFFDA}-67.33           & 69.13                         & \cellcolor[HTML]{DAFFDA}+4.27          & 55.26                         & \cellcolor[HTML]{F6FFF6}+6.54           & 90.51                                  & \cellcolor[HTML]{F6FFF6}+0.19          & 63.67                                  & \cellcolor[HTML]{FFDCDC}-0.05          & 69.64                            & \cellcolor[HTML]{F6FFF6}+2.74          \\
                                  & (13)       & \textbf{base MTL + MTD + IAS~\cite{jean2019adaptive_scheduling}}         & \cellcolor[HTML]{DAFFDA}-67.33           & \textbf{70.11}                & \cellcolor[HTML]{DAFFDA}\textbf{+5.75} & 57.97                         & \cellcolor[HTML]{C5FFC5}+11.76          & 90.88                                  & \cellcolor[HTML]{F6FFF6}+0.60          & \textbf{65.66}                         & \cellcolor[HTML]{C5FFC5}\textbf{+3.08} & 71.16                            & \cellcolor[HTML]{C5FFC5}+5.30          \\
                                  & (14)       & \textbf{base MTL + IMTLG~\cite{liu2020imtl}}                  & \cellcolor[HTML]{DAFFDA}-67.33           & 64.11                         & \cellcolor[HTML]{FF9696}-3.30          & 47.12                         & \cellcolor[HTML]{FF9696}-9.16           & 90.21                                  & \cellcolor[HTML]{FFDCDC}-0.14          & 55.61                                  & \cellcolor[HTML]{FF9696}-12.70         & 64.26                            & \cellcolor[HTML]{FF9696}-6.33          \\
\multirow{-7}{*}{\makecell{\rom{3}\\(Sec.~\ref{sec:mtd_comparison})}}             & (15)       & \textbf{base MTL + MTD + IMTLG~\cite{liu2020imtl}}            & \cellcolor[HTML]{DAFFDA}-67.33           & 67.14                         & \cellcolor[HTML]{F6FFF6}+1.27          & 57.22                         & \cellcolor[HTML]{C5FFC5}+10.31          & 90.09                                  & \cellcolor[HTML]{FFDCDC}-0.28          & 58.14                                  & \cellcolor[HTML]{FF9696}-9.56          & 68.15                            & \cellcolor[HTML]{F6FFF6}+0.44          \\ \midrule
                                  & (16)       & \textbf{\modelname (bottleneck dim. = 128)}       & \cellcolor[HTML]{F6FFF6}\textbf{-65.14}  & 70.73                         & \cellcolor[HTML]{C5FFC5}+6.68          & 58.03                         & \cellcolor[HTML]{C5FFC5}+11.88          & \textbf{91.54}                         & \cellcolor[HTML]{C5FFC5}\textbf{+1.33} & 66.20                                  & \cellcolor[HTML]{C5FFC5}+3.92          & 71.63                            & \cellcolor[HTML]{C5FFC5}+5.95          \\
                                  & (17)       & \textbf{\modelname (bottleneck dim. = 256)}       & \cellcolor[HTML]{F6FFF6}-62.67           & 71.77                         & \cellcolor[HTML]{C5FFC5}+8.25          & 58.45                         & \cellcolor[HTML]{C5FFC5}+12.69          & 91.10                                  & \cellcolor[HTML]{F6FFF6}+0.84          & 66.81                                  & \cellcolor[HTML]{C5FFC5}+4.88          & 72.03                            & \cellcolor[HTML]{C5FFC5}+6.67          \\
\multirow{-3}{*}{\makecell{\rom{4}\\(Sec.~\ref{sec:further_analysis})}}             & (18)       & \textbf{\modelname (bottleneck dim. = 512)}       & \cellcolor[HTML]{F6FFF6}-57.73           & \textbf{72.32}                & \cellcolor[HTML]{C5FFC5}\textbf{+9.08} & \textbf{58.51}                & \cellcolor[HTML]{C5FFC5}\textbf{+12.80} & 90.96                                  & \cellcolor[HTML]{F6FFF6}+0.69          & \textbf{66.92}                         & \cellcolor[HTML]{C5FFC5}\textbf{+5.05} & \textbf{72.18}                   & \cellcolor[HTML]{C5FFC5}\textbf{+6.91} \\ \bottomrule[1.5pt]
\end{tabular}%
}
\caption{
Ablation study and further analysis of our method. \texttt{Groups~(\rom{1}) and (\rom{2})}: Ablation experiments of the proposed XAA and TSA under the single-task learning (STL) and multi-task learning (MTL) scenarios.
\texttt{Group~(\rom{3})}: The comparison among our multi-teacher distillation (MTD) and other alternatives designed for task-sampling based MTL (TS-MTL).
\texttt{Group~(\rom{4})}: 
The effect of the bottleneck dimension of XAA and TSA.
\textcolor{Dandelion}{Yellow background}: The baseline performance used per column;
\textcolor{Maroon}{Red background}: negative transfer;
\textcolor{ForestGreen}{Green background}:
positive transfer.
\textbf{Bold number}: The best result in each group.
}
\label{tab:ablation_study}
\end{table*}
\statement{SCR evaluation.}
We report the performance of SCR in the left part of Tab.~\ref{tab:scr_cap_results}, following the common protocol~\cite{gao2020fashionbert,zhuge2021kaleido,han2022fashionvil}.
Similar to TGIR, our \modelname surpasses clearly all previous works~\cite{gao2020fashionbert,zhuge2021kaleido,han2022fashionvil,li2020oscar,ji2022mvlt} with heavier fusion mechanisms (\eg, modality-agnostic self-attention implemented by concatenating text tokens and image patches at the very beginning).
This validates the efficacy of our proposed XAA, 
suggesting the superiority of modality interaction 
over the conventional fusion at the input point.

\statement{FIC evaluation.}
The original FashionViL \cite{han2022fashionvil} has no decoder and cannot support generation tasks.
For comparison, we equip it with masked language modelling (MLM) auto-regressively~\cite{zhou2020vlp,li2020oscar,zhuge2021kaleido} enabling the image captioning.
The results of FIC are shown in the right part of Tab.~\ref{tab:scr_cap_results}, following the common protocol~\cite{zhuge2021kaleido}.
Our \modelname again achieves state-of-the-art performance with a clear margin.

All the above comparisons show the superior generalization capability of our method in both generative and discriminative tasks.

\subsection{Ablation study on architecture}
\label{sec:ablation_study}
Given the strong performance of our method as evaluated in Sec. \ref{sec:SOTA}, we ablate the proposed model architecture with a focus on two newly introduced adapters (TSA and XAA) in both STL and MTL settings.

\statement{Single-task learning setting.}
For comparison on XMR, TGIR and SCR, we design the \texttt{baseline}
as directly fine-tuning the vanilla CLIP without any new modules (L1). 
With the two-stream design, CLIP cannot tackle FIC, and we hence further equip it with our XAA as the \texttt{baseline} for FIC (L3).
From the results shown in group~(\rom{1}) of Tab.~\ref{tab:ablation_study}, we find that TSA and XAA can bring in 3\%-6\% relative improvements for XMR and TGIR.
In particular, XAA gives TGIR a significant improvement, demonstrating the superiority of our layer-wise modality interaction mechanism.
However, these adapters have only  a marginal impact on the performance of SCR and FIC, with a performance drop of less than 0.5\% when the model is independently trained on a single task.

\statement{Multi-task learning setting.}
Similarly, we construct the \texttt{baselines} for the MTL setting using the vanilla CLIP and XAA-equipped CLIP (L5 and L7).
As shown in L5 in the group~(\rom{2}) of Tab.~\ref{tab:ablation_study},
a severe negative transfer occurs with an overall 5.59\% performance drop.
Likewise, there is also a negative transfer for the XAA-equipped CLIP model (L7),
albeit with a slight increase in performance.
This suggests the challenges of heterogeneous multi-task learning in the fashion domain.
This problem can be well solved using our TSA,
with an overall 4\%$\sim$6\% improvement (L5 \vs L7 and L6 \vs L8), even though only a few extra task-specific parameters are introduced (1.35\% of the original CLIP size).
Interestingly, we also found that XAA and TSA are reciprocal:
\textbf{(1)} When TSA and XAA work together, the model can achieve better relative performance than the sum of their gains (L4 \vs L2+L3 and L8 \vs L6+L7)
%
\textbf{(2)} When TSA or XAA is applied in isolation, the multi-task model always underperforms its single-task counterpart (L6 \vs L2 and L7 \vs L3).
But the multi-task model with both TSA and XAA exceeds the single-task counterpart (L8 vs L4), indicating that TSA and XAA play complementary roles better in the multi-task setting,
as expected and designed so.

\subsection{Ablation study on multi-task training strategy}
\label{sec:mtd_comparison}
Following the above architecture analysis, 
we further ablate the proposed multi-teacher distillation (MTD) based training strategy. 
We compare extensively with previous sampling strategies and gradient manipulation algorithms.

\statement{Task sampling.}
We start by comparing two common sampling strategies (\texttt{uniform} and \texttt{round-robin}) with our \texttt{size-proportional} strategy.
Round-robin sampling is a special case of uniform sampling -- each task is sampled one by one.
As shown in L9-L11 in the group~(\rom{3}) of Tab.~\ref{tab:ablation_study}, both uniform and round-robin sampling underperform our size-proportional sampling by a gap of 1.5\%.
This is due to imbalanced dataset sizes across different tasks, which is ignored in uniform sampling and round-robin sampling.

\statement{Gradient manipulation.}
To compare with our MTD scheme, we consider two kinds of gradient manipulation algorithms: \texttt{Implicit Adaptive Scheduling (IAS)}~\cite{jean2019adaptive_scheduling} and  \texttt{IMTLG}~\cite{liu2020imtl}.
In particular, IAS is a representative strong method that adaptively changes the task sampling ratio, learning rate, or gradient scale for each task~\cite{wang2020makes,lu202012-in-1}.
Specifically, it scales the gradients of each task according to the performance on the validation set.
Instead, IMTLG is a representative of those methods manipulating all the gradients together
\cite{chen2018gradnorm,sener2018mgda,kendall2018uncertainty,navon2022nash-mtl}.
It is featured by a closed-form solution to optimize the scaling factors of each task, such that the aggregated gradients (sum of raw gradients weighted by the scaling factors) have equal projections onto individual tasks.
Since IMTLG cannot be directly applied to task-sampling based MTL, we further adapt it by maintaining a gradient buffer to store the gradients of each task and update the parameters every four iterations (each corresponding one of the four fashion tasks).
As shown in the group~(\rom{3}) of Tab.~\ref{tab:ablation_study}, the performance of IAS (L12) and IMTLG (L14) are significantly lower than that of our MTD (L9). In particular, IMTLG suffers from a severe negative transfer (-6.33\%).
There are two plausible reasons:
\textbf{(1)} Relying on a heuristic strategy, IAS struggles in
finding the optimal status over all tasks, despite the access to the validation performance.
%
\textbf{(2)} IMTLG may experience over-fitting for the tasks with smaller training data (\eg, TGIR), which cannot be addressed by the idea of ensuring the final gradient direction to have the same impact on each task. 
On the contrary, our MTD can implicitly regularize the gradients via knowledge distillation, without a costly need of monitoring the validation performance.
Guided by the soft ground truth of each teacher,
overfitting can be well avoided in an elegant manner.
Considering the methodical orthogonality, we further apply our MTD on top of IAS (L13) and IMTLG (L15).
It is shown that this can improve both by a large margin (L13 \vs L12 and L15 \vs L14), demonstrating 
the generic usability of our training method.

\begin{figure}[t]
\begin{center}
\includegraphics[width=\linewidth]{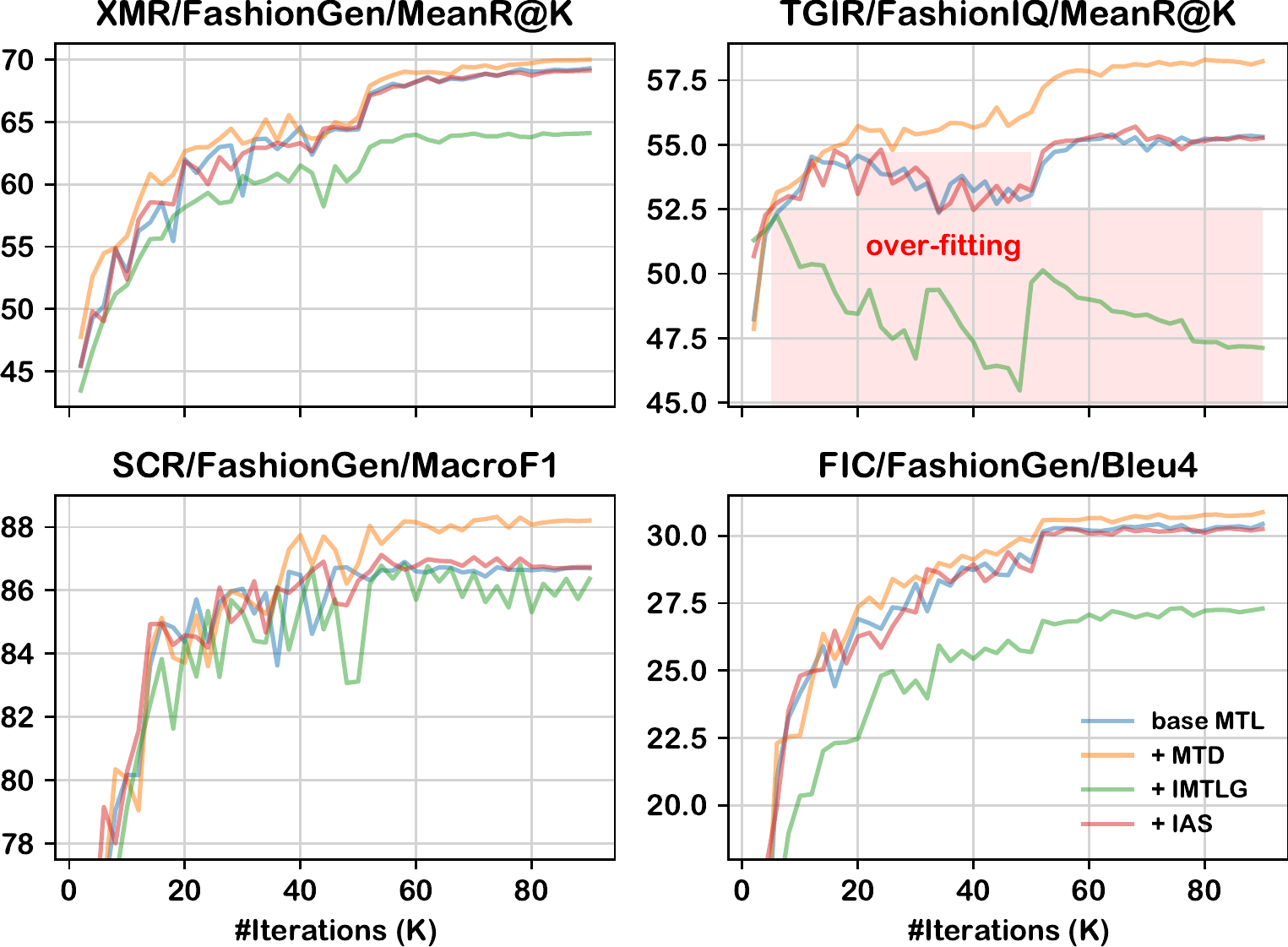}
\end{center}
\caption{Training dynamics of our multi-teacher distillation (MTD) and alternative multi-task learning methods (IAS \cite{jean2019adaptive_scheduling} and IMTLG \cite{liu2020imtl}).
Metric: The validation performance curves. }
\label{fig:curves}
\end{figure}
\subsection{Further analysis}
\label{sec:further_analysis}
\statement{Regularizing effect of MTD.}
To shed more light on the regularization effect of MTD, we plot the validation performance curves in Fig.~\ref{fig:curves}.
Without MTD, the baseline MTL model
is prone to overfit on TGIR after about $20k$ iterations
due to much less training data than other tasks.
Interestingly, this overfitting is even amplified by IMTLG. 
This is because IMTLG needs to pay more attention to TIRG in order to achieve impartial learning.
Overall, neither IAS nor IMTLG can improve over the baseline MTL, regardless of overfitting or not.
Encouragingly, our MTD yields consistent and
significant performance boost per task, 
rendering it a more stable and effective learning strategy.

\statement{Scaling up bottleneck dimension.}
We evaluate the effect of the bottleneck dimension of the AdaptMLP in XAA and TSA (the only hyper-parameter of our architecture).
We vary this dimension from $64$ to $512$.
As shown in the group~(\rom{4}) of Tab.~\ref{tab:ablation_study}, it is evident that the overall relative performance is positively correlated with this bottleneck dimension.
This indicates that \modelname could be potentially more performing at the cost of more parameters.
Also, we observe a trade-off between model size increase  and performance gain. 
For example, 10\% more parameters are needed for exchanging a relative performance gain of 1.4\% (L18 \vs L9). 
We also notice that further improvement is not consistent over tasks. For instance, the performance of SCR will gradually deteriorate with the increase of bottleneck dimension.
An interesting direction for future works could be exploiting adaptive algorithms~\cite{sun2020adashare,sun2022gppf} to optimize the best bottleneck dimension per task.

\statement{Qualitative examples.}
To reflect the output of \modelname more intuitively, in addition to Fig.~\ref{fig:tasks}, we show more illustrative outputs from \modelname in the supplementary file.

\section{Conclusions}
We have introduced \modelname for heterogeneous fashion tasks, grounded upon a generic off-the-shelf V+L model.
It addresses cross-modal retrieval, text-guided image retrieval, multi-modal classification, and image captioning in a unified architecture.
This is made possible by the proposed task-versatile architecture with cross-attention adapters and task-specific adapters, and a scalable multi-task training pipeline with multi-teacher distillation. 
Extensive experiments showed that our \modelname achieves new state-of-the-art performance on all tasks with significantly fewer parameters.
%
\appendix
\section{Implementation details}
\begin{table}[t]
\centering
\resizebox{\linewidth}{!}{
\begin{tabular}{llc}
\toprule[1.5pt]
\multirow{3}{*}{\textbf{Model architecture}}& Vision encoder (VE)  & CLIP (ViT-B/16)~\cite{radford2021clip}   \\
                                            & Language encoder (LE)& CLIP (ViT-B/16)~\cite{radford2021clip}   \\
                                            & Bottleneck dim. & 64                 \\ \midrule
\multirow{3}{*}{\textbf{Data augmentation}} & Resize               & (256, 256)         \\
                                            & RandomCrop           & (224, 224)         \\
                                            & RandomHorizontalFlip & \checkmark                \\ \midrule
\multirow{11}{*}{\textbf{Training setting}} & Number of iterations & $90k$             \\
                                            & Batch size           & 64                 \\
                                            & Initial LR of VE/LE  & 1e-6               \\
                                            & Initial LR of Adapters    & 1e-4               \\
                                            & LR schedule          & Multi-step         \\
                                            & LR steps             & $50k$ and $80k$  \\
                                            & LR decrease ratio    & 0.1                \\
                                            & Warmup iterations    & $10k$             \\
                                            & Warmup factor        & 0.25               \\
                                            & Optimizer            & AdamW (0.9, 0.999) \\
                                            & Weight decay         & 1e-5               \\ \midrule
\multirow{2}{*}{\textbf{Hardware}}          & GPU                  & 4 $\times$ RTX 3090       \\
                                            & Training duration    & 31.5h              \\ \bottomrule[1.5pt]
\end{tabular}
}
\caption{Details for multi-task training \modelname.}
\label{tab:training_hyper}
\end{table}
This section describes our implementation and multi-task training details for \modelname.

\statement{Architecture details.} 
As mentioned in the main paper, we build our \modelname upon off-the-shelf CLIP model~\cite{radford2021clip}.
We utilize the ViT-B/16 version and get the pre-trained weights from HuggingFace Transformers~\cite{wolf2020huggingface}.
Specifically, as described in the original paper~\cite{radford2021clip}, the language encoder is a 12-layer 512-wide Transformer~\cite{vaswani2017transformer} with 8 attention heads, while the vision encoder is a base-size Vision Transformer (ViT)~\cite{dosovitskiy2020vit} with patch size as 16.
Masked self-attention was used in the language encoder.
For computational efficiency, the max text sequence length was capped at 76.
The text sequence is bracketed with \texttt{[SOS]} and \texttt{[EOS]} tokens and the activations of the highest layer of the language encoder at the \texttt{[EOS]} token are treated as the text feature representation.
Please find more details about CLIP and its pre-training in the original paper~\cite{radford2021clip}.

\begin{figure}[t]
\centering
\captionsetup{position=top}
\subfloat[\textbf{Text query:} Satin cap in black. Adjustable snapback fastening. Tonal hardware. Tonal stitching.]{
\includegraphics[width=\linewidth]{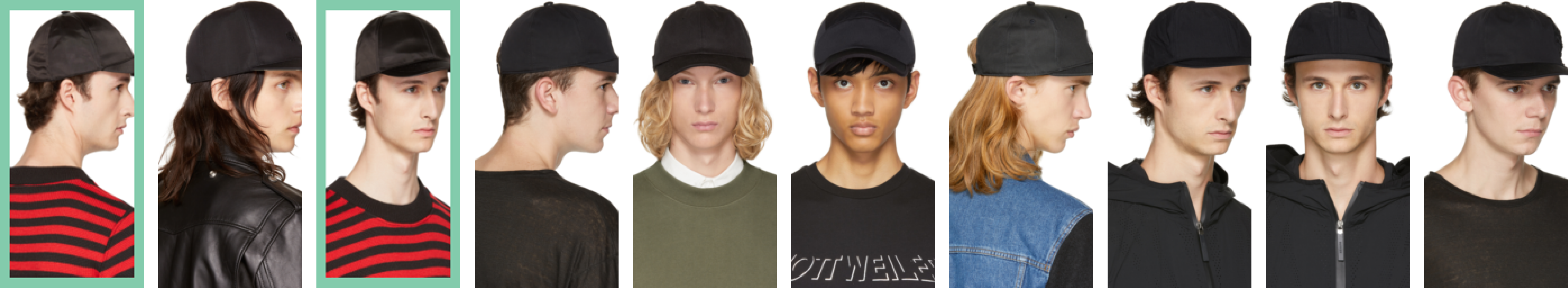}
}
\vspace{2mm}
\subfloat[\textbf{Text query:} French terry lounge shorts in marled grey. Elasticized waistband. Three-pocket styling. Zip-fly.]{
\includegraphics[width=\linewidth]{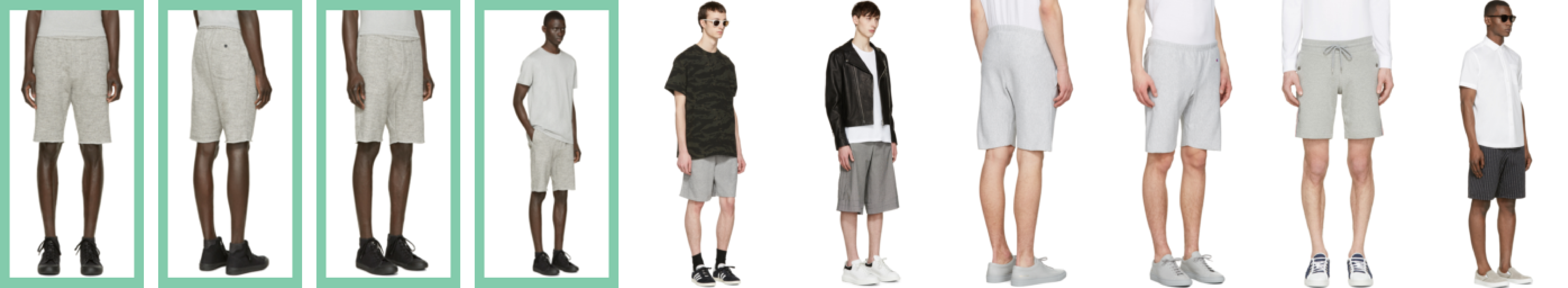}
}
\vspace{2mm}
\subfloat[\textbf{Text query:} Wide-leg woven cotton sarouel-style trousers in dark navy. Partially elasticized waistband. Pleats at front. Two-pocket styling. Unlined.]{
\includegraphics[width=\linewidth]{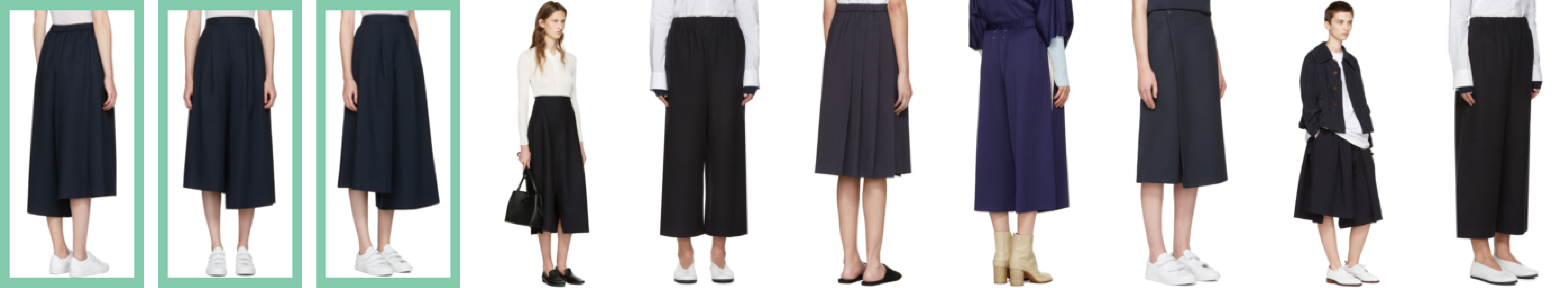}
}
\vspace{2mm}
\subfloat[\textbf{Text query:} Relaxed-fit sweatshirt in heather grey. Ribbed knit crewneck collar, cuffs, and hem. Raglan sleeves. Mock calf hair at breast in red.]{
\includegraphics[width=\linewidth]{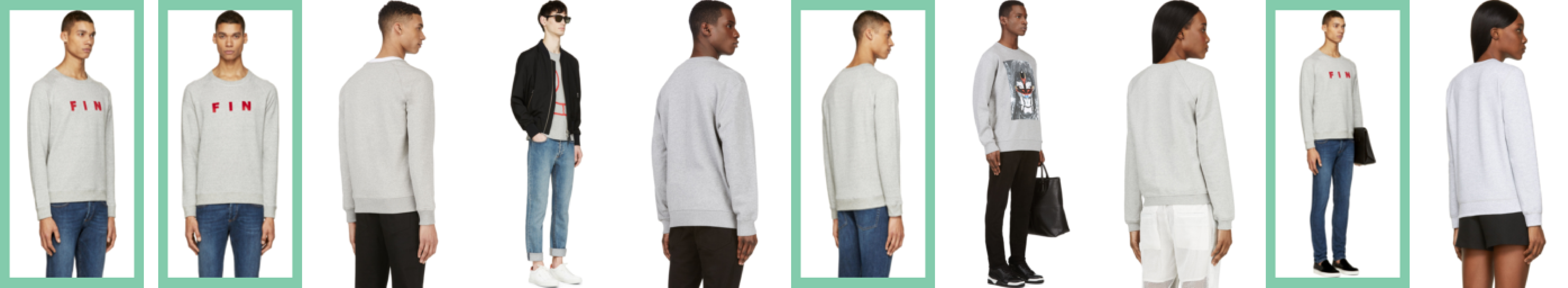}
}
\vspace{2mm}
\subfloat[\textbf{Text query:} Short sleeve t-shirt in black. Rib-knit crew-neck collar. Logo printed at front in white and black. Tonal stitching.]{
\includegraphics[width=\linewidth]{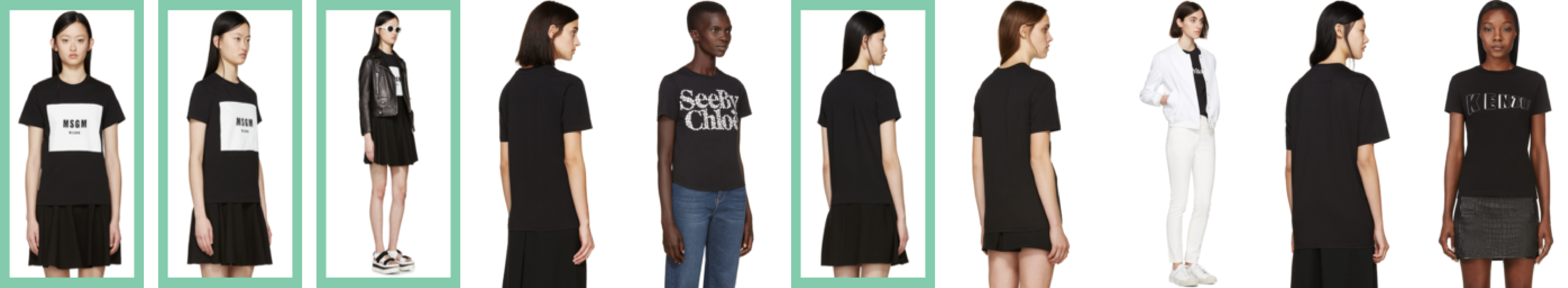}
}
\vspace{4mm}
\caption{XMR examples. \textcolor{ForestGreen}{Green box} indicates the ground truth.}
\label{fig:more_xmr}
\end{figure}
\statement{Training details.}
We list all hyper-parameters used for multi-task training in 
 Tab.~\ref{tab:training_hyper}, including data augmentation methods, optimizer setting, scheduler setting, and \etc.
 This results in about 31.5 hours of training time on four RTX 3090 GPUs (24GB memory for each).
 For single-task training (used by single-task teachers training and ablation study), we adopted the same hyper-parameters except for shorter training iterations ($30k$ for tasks on FashionGen~\cite{rostamzadeh2018fashiongen}, $6k$ for tasks on FashionIQ~\cite{wu2021fashioniq}).

\begin{table}[t]
\centering
\resizebox{\columnwidth}{!}{%
\begin{tabular}{cccccc}
\toprule[1.5pt]
\multicolumn{3}{c}{\textbf{XMR* (Image to Text)}}                       & \multicolumn{3}{c}{\textbf{XMR* (Text to Image)}}              \\
\textbf{R@1}  & \textbf{R@5}                     & \textbf{R@10}       & \textbf{R@1}        & \textbf{R@5}         & \textbf{R@10}                                      \\ \midrule
 $45.99\pm0.25$      &                  $73.25\pm0.15$      &  
 $81.84\pm0.12$      & 
 $53.12\pm0.07$      &  
 $77.55\pm0.35$      &   
 $86.02\pm0.22$       \\ \midrule[1.5pt]
\multicolumn{2}{c}{\textbf{TGIR (Dress)}}        & \multicolumn{2}{c}{\textbf{TGIR (Shirt)}} & \multicolumn{2}{c}{\textbf{TGIR (Toptee)}}  \\
\textbf{R@10} & \textbf{R@50}                    & \textbf{R@10}       & \textbf{R@50}       & \textbf{R@10}        & \textbf{R@50}                                     \\ \midrule
$42.16\pm0.32$   & 
$67.10\pm0.22$   & 
$47.19\pm0.33$   &
$67.91\pm0.66$   &  
$50.79\pm0.08$   &       
$73.48\pm0.38$                                 \\ \midrule[1.5pt]
\textbf{SCR (Acc)}  & \multicolumn{1}{c|}{\textbf{SCR (F1)}} & \textbf{FIC (B)}          & \textbf{FIC (M)}          & \textbf{FIC (R)}           & \textbf{FIC (C)}                                         \\ \midrule
              $94.62\pm0.08$ & 
              \multicolumn{1}{c|}{$87.80\pm0.25$}            &     $30.59\pm0.15$       &       $24.22\pm1.99$       &
              $55.69\pm0.13$       &                   $148.9\pm1.19$                               \\ \bottomrule[1.5pt]
\end{tabular}%
}
\caption{Statistical significance quantification of our results.
* XMR evaluation is under \texttt{full database} protocol
}
\label{tab:condidence}
\end{table}
\section{Additional quantitative results}
We followed the same protocol used by previous works~\cite{han2022fashionvil} and used the same random seed for training, to ensure a direct comparison to these main competitors. 
We also trained our model two more times with different random seeds to measure our method's stability.
The statistical performance (w/ mean and std) over three trials in Tab.~\ref{tab:condidence} shows that our model is stable. 

\section{Additional qualitative results}
\begin{figure}[t]
\centering
\captionsetup{position=top}
\subfloat[\textbf{Modifying text:} the shirt is purple and black, has slightly longer sleeves and is purple and black.]{
\includegraphics[width=\linewidth]{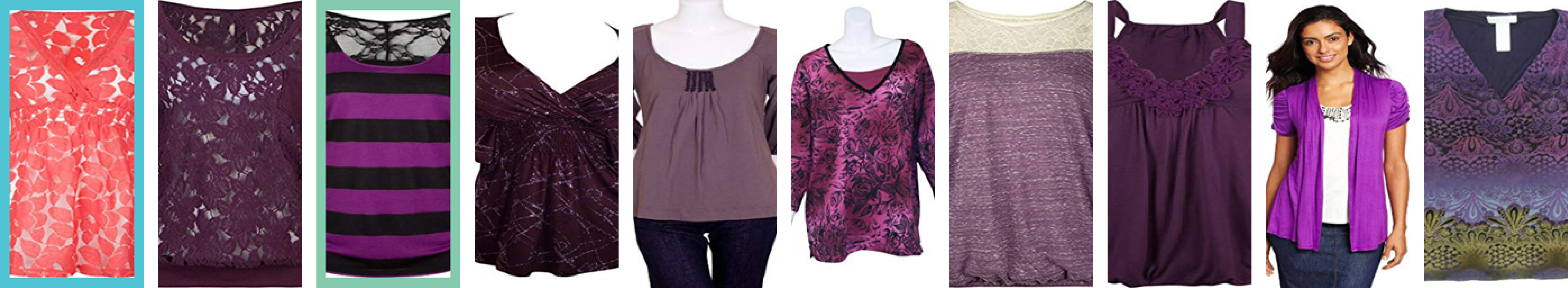}
}
\vspace{2mm}
\subfloat[\textbf{Modifying text:} is a green t-shirt with a light material, is more colorful.]{
\includegraphics[width=\linewidth]{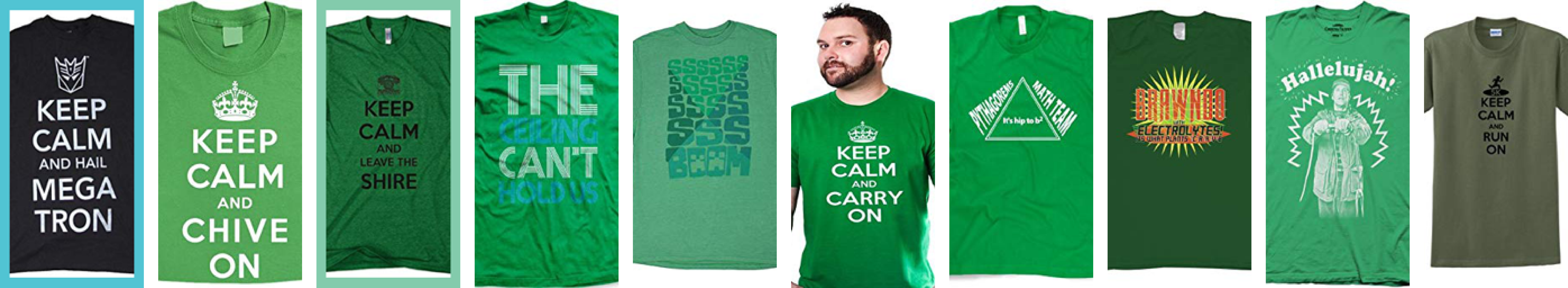}
}
\vspace{2mm}
\subfloat[\textbf{Modifying text:} is blue with a collar and some buttons, is blue and shorter sleeved.]{
\includegraphics[width=\linewidth]{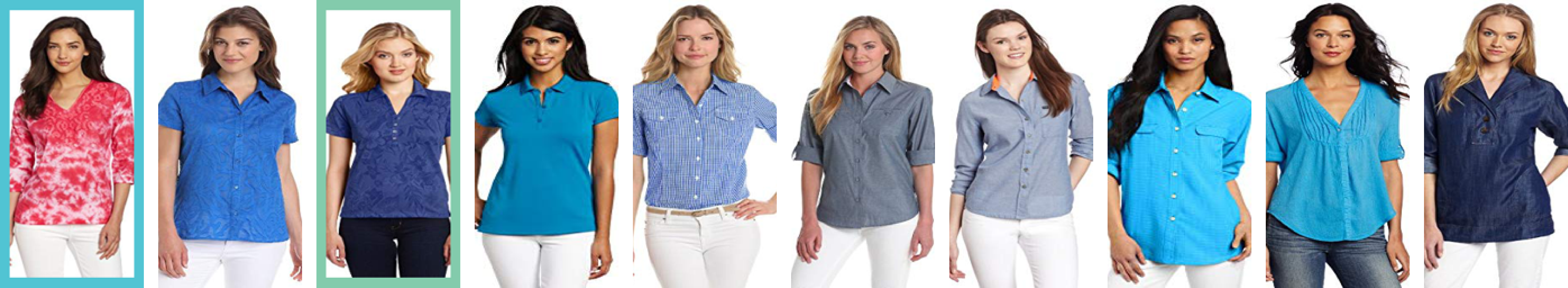}
}
\vspace{2mm}
\subfloat[\textbf{Modifying text:} is maroon with a ruffled top, is a dark red cowl-neck and long sleeves.]{
\includegraphics[width=\linewidth]{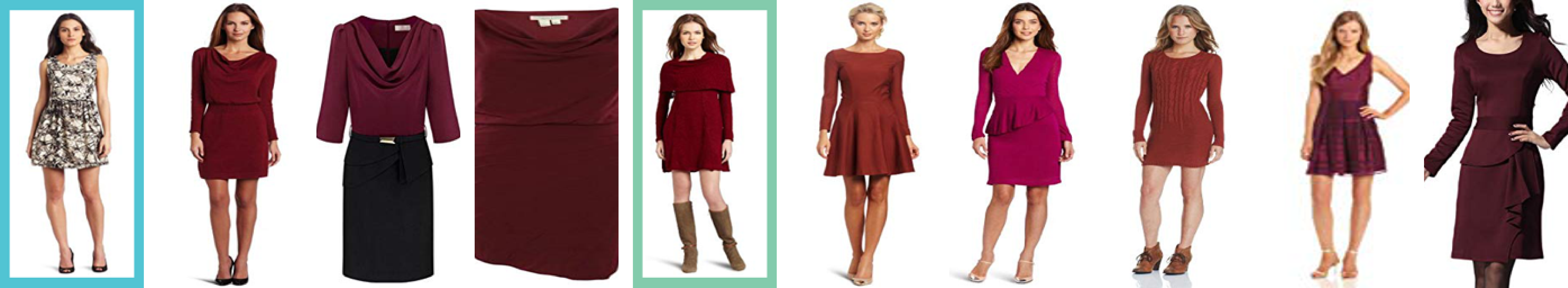}
}
\vspace{2mm}
\subfloat[\textbf{Modifying text:} is more plain and has tank top sleeves, is shorter and souped neck.]{
\includegraphics[width=\linewidth]{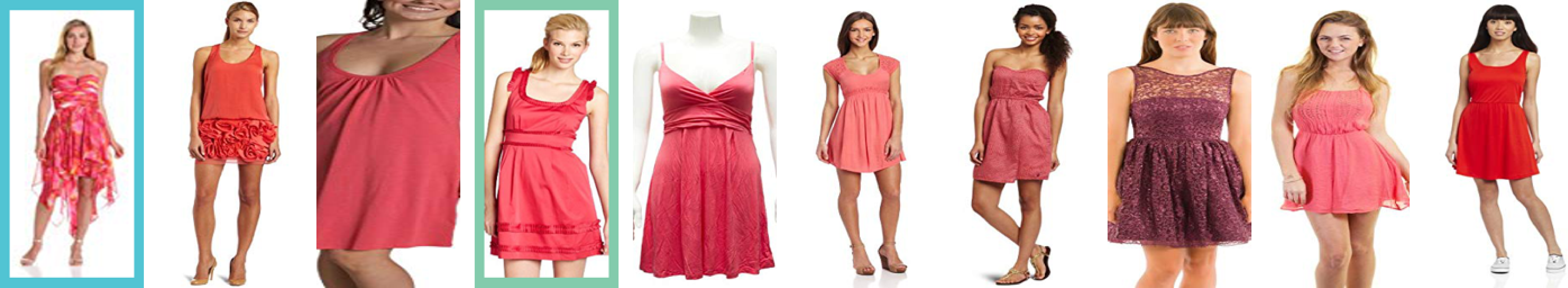}
}
\vspace{4mm}
\caption{TGIR examples. \textcolor{Turquoise}{Blue box} indicates the reference image. \textcolor{ForestGreen}{Green box} indicates the ground truth.}
\label{fig:more_tgir}
\end{figure}
\begin{table}[t]
\centering
\resizebox{\linewidth}{!}{%
\begin{tabular}{ccm{4cm}m{3.2cm}}
\toprule[1.5pt]
& \textbf{Images} & \textbf{Ground Truth Captions} & \textbf{Generated Captions} \\ \midrule
(a)
& $\raisebox{-.5\height}{\includegraphics[width=0.2\linewidth]{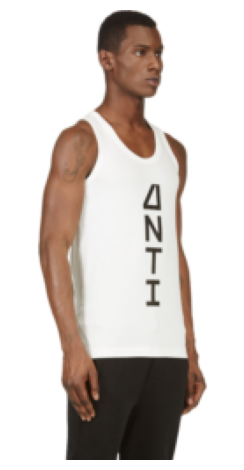}}$                
& \textcolor{ForestGreen}{White logo tank top}. Relaxed-fit \textcolor{ForestGreen}{tank top in white}. Ribbed \textcolor{ForestGreen}{scoopneck collar} and armscyes. \textcolor{ForestGreen}{Logo print at black}. Tonal logo embroidered at back hem. \textcolor{ForestGreen}{Tonal stitching}.                 
& \textcolor{ForestGreen}{White logo tank top}. Racer-back \textcolor{ForestGreen}{tank top in white}. \textcolor{ForestGreen}{Scoopneck collar}. \textcolor{ForestGreen}{Logo printed} at front \textcolor{ForestGreen}{in black}. Curved hem. \textcolor{ForestGreen}{Tonal stitching}. 
\\\midrule
(b)
& $\raisebox{-.5\height}{\includegraphics[width=0.2\linewidth]{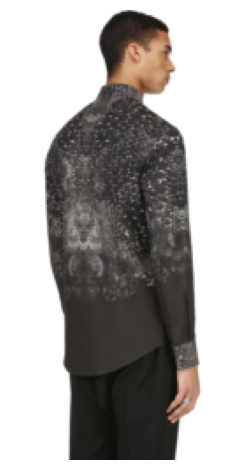}}$
& \textcolor{ForestGreen}{Black} python print \textcolor{ForestGreen}{shirt}. Short sleeve shirt in tones of grey and \textcolor{ForestGreen}{black}. Detailed python scale print throughout with ombre effect at bottom portions. \textcolor{ForestGreen}{Spread collar. Button closure at front. Tonal stitching. Single-button barrel cuffs with buttoned sleeve placket.}                    
& \textcolor{ForestGreen}{Black} paint splatter \textcolor{ForestGreen}{shirt}. Long sleeve shirt in \textcolor{ForestGreen}{black}. Graphic print throughout in white. \textcolor{ForestGreen}{Spread collar. Button closure at front. Tonal stitching. Single-button barrel cuffs with buttoned sleeve placket.} 
\\\midrule
(c)
& $\raisebox{-.5\height}{\includegraphics[width=0.2\linewidth]{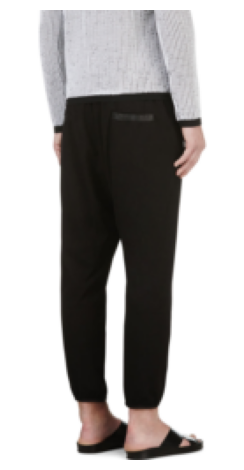}}$
& \textcolor{ForestGreen}{Black} jersey leather trim \textcolor{ForestGreen}{lounge pants}. Leather-trimmed stretch jersey \textcolor{ForestGreen}{lounge pants in black}. Partially \textcolor{ForestGreen}{elasticized waistband} with leather \textcolor{ForestGreen}{ drawstring closure}. \textcolor{ForestGreen}{Zip fly}. Leather pocket trim. Elasticized grosgrain cuffs.
& \textcolor{ForestGreen}{Black lounge pants}. \textcolor{ForestGreen}{Lounge pants in black}. \textcolor{ForestGreen}{Elasticised waistband with drawstring closure}. Four-pocket styling. Elasticised ankle cuffs. Tonal stitching. \textcolor{ForestGreen}{Zip fly}.
\\\midrule
(d)
& $\raisebox{-.5\height}{\includegraphics[width=0.2\linewidth]{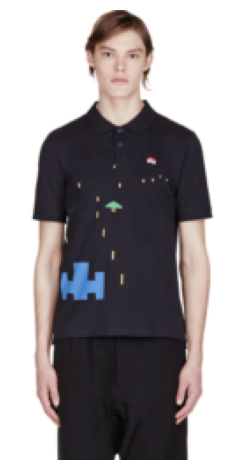}}$
& \textcolor{ForestGreen}{Navy} pixel print atari edition \textcolor{ForestGreen}{polo}. \textcolor{ForestGreen}{Short sleeve} oversized \textcolor{ForestGreen}{polo in navy}. Atari pixel print at front. \textcolor{ForestGreen}{Spread collar} with two\textcolor{ForestGreen}{-button placket}. Slits at side seams. \textcolor{ForestGreen}{Tonal stitching}.
& \textcolor{ForestGreen}{Navy} embroidered patch \textcolor{ForestGreen}{polo}. \textcolor{ForestGreen}{Short sleeve} cotton piqu \& eacute \textcolor{ForestGreen}{polo in navy}. Ribbed \textcolor{ForestGreen}{spread collar} and trim at sleeve opening. Five\textcolor{ForestGreen}{-button placket} at front. Signature tri-color tab at back collar. Tennis tail hem. \textcolor{ForestGreen}{Tonal stitching}.
\\\midrule
(e)
& $\raisebox{-.5\height}{\includegraphics[width=0.2\linewidth]{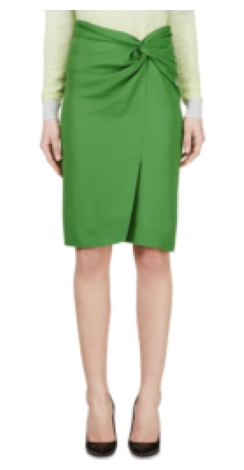}}$
& \textcolor{ForestGreen}{Green} wrap pencil \textcolor{ForestGreen}{skirt}. High-waisted wrap pencil \textcolor{ForestGreen}{skirt in green}. Gathered knot detail at waist. Vent at front hem. Zip closure at back. \textcolor{ForestGreen}{Tonal stitching}.
& \textcolor{ForestGreen}{Green} silk draping \textcolor{ForestGreen}{skirt}. Silk \textcolor{ForestGreen}{skirt in green}. Elasticized waistband with drawstring at interior. Vented at back waist. Seam pockets at sides. \textcolor{ForestGreen}{Tonal stitching}.
\\ \bottomrule[1.5pt]
\end{tabular}%
}
\caption{FIC examples. \textcolor{ForestGreen}{Green text} indicates the matched phrases.}
\label{fig:more_fic}
\end{table}
\begin{figure}[t]
\centering
\captionsetup{position=top}
\subfloat[\textbf{Text query:} Skinny-fit stretch denim jeans in 'fade to' grey. Fading and whiskering throughout. Mid-rise. Five-pocket styling.]{
\includegraphics[width=\linewidth]{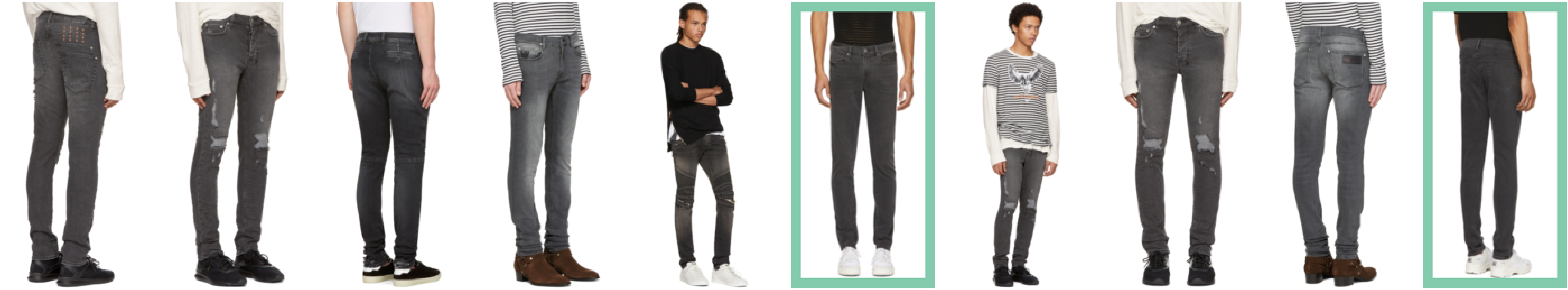}
}
\vspace{2mm}
\subfloat[\textbf{Modify text:} is beige and sleeveless, is blue t-shirt with owls on front.]{
\includegraphics[width=\linewidth]{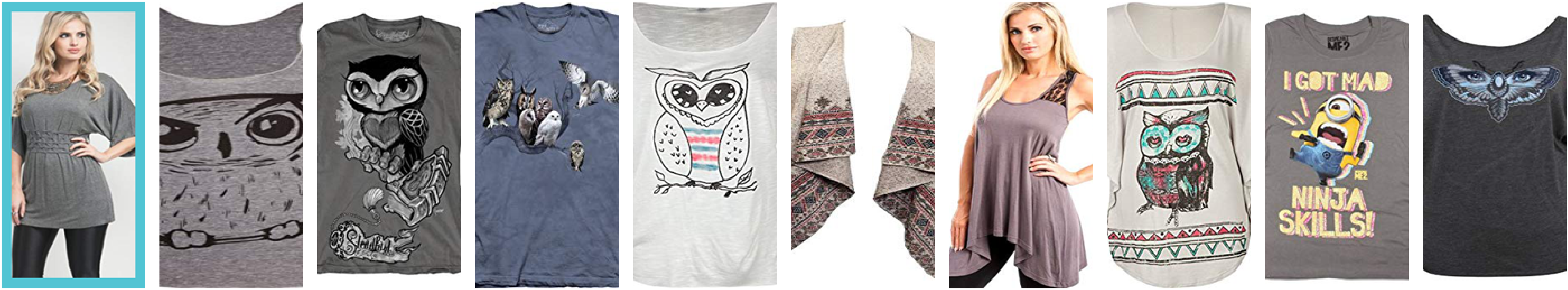}
}
\vspace{4mm}
\caption{Failure cases. \textcolor{Turquoise}{Blue box} indicates the reference image. \textcolor{ForestGreen}{Green box} indicates the ground truth.}
\label{fig:failure_case}
\end{figure}

\begin{figure}[t]
\centering
\includegraphics[width=\linewidth]{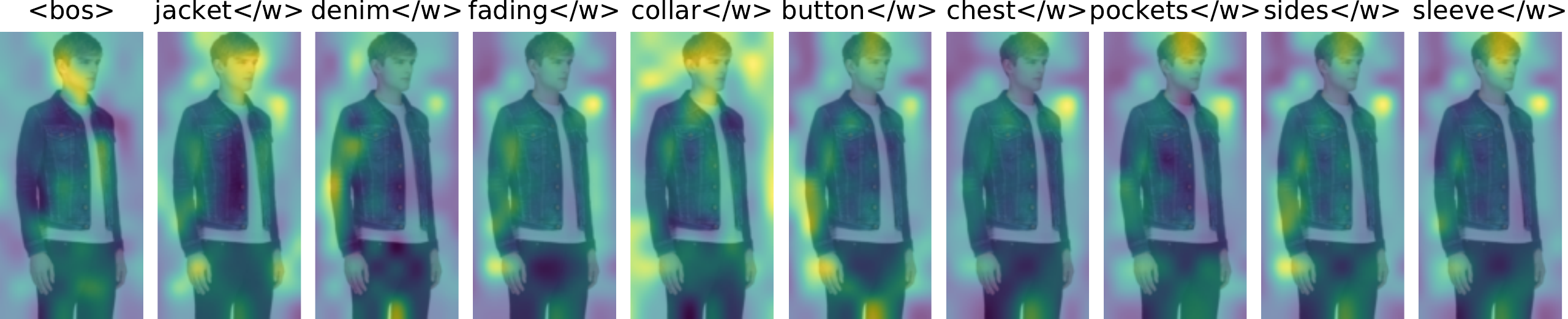}
\includegraphics[width=\linewidth]{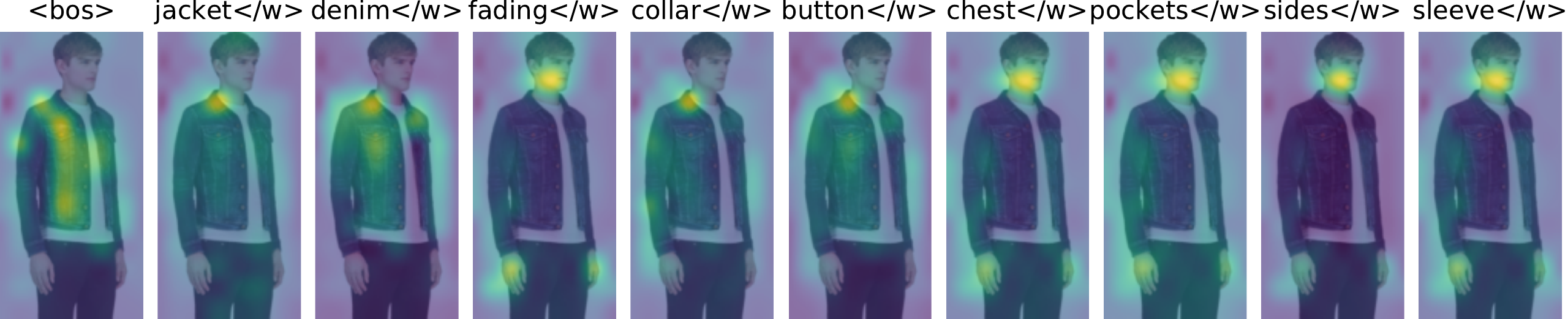}
\caption{Visualized attention maps of SCR.}
\label{fig:attn}
\end{figure}

We provide more visualization results in this section to better understand the performance of our \modelname in a qualitative way.
Specifically, we show cross-modal retrieval (XMR) results in Fig.~\ref{fig:more_xmr}, text-guided image retrieval (TGIR) results in Fig.~\ref{fig:more_tgir} and fashion image captioning (FIC) results in Tab.~\ref{fig:more_fic}.
We didn't show subcategory recognition (SCR) results here because of the lack of intuition when visualizing this classification task, but the visualized attention maps are given in Fig.~\ref{fig:attn}.

For retrieval tasks (XMR and TGIR), we observe ambiguities (\ie, the ground truth is not the only one matching the query) in the fashion datasets~\cite{wu2021fashioniq,rostamzadeh2018fashiongen}.
Especially in FashionIQ, there are many false negatives that are neglected during the data-annotation stage.
Even so, our \modelname can offer us a reliable and human-understandable ranking list, demonstrating its superiority in fine-grained discrimination. 
Fig.~\ref{fig:failure_case} shows example failure cases from (a) XMR and (b) TGIR. 
In the text query example, we can see that even the human-annotated ground truth (indicated by green boxes) images do not fit the text query perfectly. In both failure cases, the top retrieved results, though wrong according to the ``ground truth'', are still largely aligned with the query/modify text.  

Because of the fine-grained nature of the fashion domain, the ground truth captions in fashion contain much more fine-grained phrases than those in the generic domain~\cite{han2022fashionvil}.
Despite this challenge, our \modelname can produce concrete and accurate phrases in the generated captions.
Even if some of the generated phrases do not exist in the ground truth, they still conform to the content of the image and human intuition.
This point proves the effectiveness of \modelname in fine-grained generation.

To gain a more intuitive understanding of how attention
is learned in our model, we visualize the \textit{text-to-image attention maps} (the average over all heads) in the \textit{last XAA} of the STL baseline (first row) and our MTL model (second row), as shown in Fig.~\ref{fig:attn}.
It is observed that the attention maps from our model are more accurate and meaningful.

{\small
\bibliographystyle{ieee_fullname}
\bibliography{egbib}
}

\end{document}